%% file: iclr2023_conference.tex
\documentclass{article} 
\usepackage{iclr2023_conference,times}
\usepackage{wrapfig}
\usepackage{booktabs} 
\usepackage[font=small]{caption}
\usepackage{hyperref}
\usepackage{url}

\input{src/imports}

\title{
Neural Constraint Satisfaction:\\
Hierarchical Abstraction for Combinatorial Generalization in Object Rearrangement}

\author{
  Michael Chang\thanks{
  UC Berkeley. 
  $^\dagger$Meta AI Research.
  $^\ddagger$Princeton University.
  $^\star$UT Austin.\newline
  Work done while MC interned at Meta AI. Correspondence to: mbchang@berkeley.edu, amyzhang@meta.com
  }, 
  Alyssa L. Dayan$^*$, Franziska Meier$^\dagger$, \\
  \textbf{ Thomas L. Griffiths$^\ddagger$, Sergey Levine$^*$, Amy Zhang$^{\dagger,\star}$} \\
}

\iclrfinalcopy 
\begin{document}

\maketitle
\input{src/abstract}
\input{src/intro}
\input{src/related_work}
\input{src/problem}
\input{src/method}
\input{src/experiments}
\input{src/discussion}
\input{src/acknowledgements}



\bibliography{bibliography}
\bibliographystyle{iclr2023_conference}

\appendix
\input{src/appendix}

\end{document}

%% file: src/imports.tex
\usepackage{amsmath}
\usepackage{graphicx}
\usepackage{mathtools}
\usepackage{esvect}
\usepackage{bbm}
\usepackage{bm}
\usepackage{lipsum}
\usepackage{xspace}
\usepackage[capitalize,noabbrev]{cleveref}
\usepackage{todonotes}
\usepackage{soul}

\usepackage{subcaption}

\usepackage{xparse}

\ExplSyntaxOn

\makeatletter
\NewDocumentCommand{\multicitep}{m}
 {
  \NAT@open
  \mjb_multicitep:n { #1 }
  \NAT@close
 }
\makeatother

\seq_new:N \l_mjb_multicite_in_seq
\seq_new:N \l_mjb_multicite_out_seq
\seq_new:N \l_mjb_cite_seq

\cs_new_protected:Npn \mjb_multicitep:n #1
 {
  \seq_set_split:Nnn \l_mjb_multicite_in_seq { ; } { #1 }
  \seq_clear:N \l_mjb_multicite_out_seq
  \seq_map_inline:Nn \l_mjb_multicite_in_seq
  {
    \mjb_cite_process:n { ##1 }
  }
  \seq_use:Nn \l_mjb_multicite_out_seq { ;~ }
 }

\cs_new_protected:Npn \mjb_cite_process:n #1
 {
  \seq_set_split:Nnn \l_mjb_cite_seq { , } { #1 }
  \int_compare:nTF { \seq_count:N \l_mjb_cite_seq == 1 }
  {
    \seq_put_right:Nn \l_mjb_multicite_out_seq
     { \citep{#1} }
  }
  {
    \seq_put_right:Nx \l_mjb_multicite_out_seq
     {
      \exp_not:N \citenum{\seq_item:Nn \l_mjb_cite_seq { 1 }},~
      \seq_item:Nn \l_mjb_cite_seq { 2 }
     }
  }
 }
\ExplSyntaxOff

\usepackage{color}

\usepackage{graphicx}

\usepackage{float}

\usepackage{morefloats}

\usepackage{placeins}

\usepackage{xspace}

\usepackage{multirow}

\usepackage{ltablex}

\usepackage{booktabs}

\newcolumntype{Y}{>{\centering\arraybackslash}X}

\usepackage{algorithm}
\usepackage{algorithmicx}
\usepackage{algpseudocode}
\usepackage{amsmath,amssymb}
\usepackage[amsmath]{ntheorem}
\usepackage{mathtools}
\usepackage{dsfont}

\DeclareMathOperator*{\argmax}{arg\,max}

\DeclarePairedDelimiterX{\infdivx}[2]{(}{)}{%
  #1\;\delimsize|\delimsize|\;#2%
}

\usepackage{optidef}

\newcommand{\methodname}{NCS\xspace}

\algnewcommand\algorithmicforeach{\textbf{for each}}
\algdef{S}[FOR]{ForEach}[1]{\algorithmicforeach\ #1\ \algorithmicdo}

%% file: src/abstract.tex
\begin{abstract}
Object rearrangement is a challenge for embodied agents because solving these tasks requires generalizing across a combinatorially large set of configurations of entities and their locations. Worse, the representations of these entities are unknown and must be inferred from sensory percepts. We present a hierarchical abstraction approach to uncover these underlying entities and achieve combinatorial generalization from unstructured visual inputs. By constructing a factorized transition graph over clusters of entity representations inferred from pixels, we show how to learn a correspondence between intervening on states of entities in the agent's model and acting on objects in the environment. We use this correspondence to develop a method for control that generalizes to different numbers and configurations of objects, which outperforms current offline deep RL methods when evaluated on simulated rearrangement tasks.
\href{https://sites.google.com/view/neural-constraint-satisfaction/home}{\textcolor{blue}{Project website}.}
\end{abstract}

%% file: src/intro.tex
\input{src/figs/two_level_hierarchy}
\section{Introduction}
The power of an abstraction depends on its usefulness for solving new problems.
Object rearrangement~\citep{batra2020rearrangement} offers an intuitive setting for studying the problem of learning reusable abstractions.
Solving novel rearrangement problems requires an agent to not only infer object representations without supervision, but also recognize that the same action for moving an object between two locations can be reused for different objects in different contexts.

We study the simplest setting in simulation with pick-and-move action primitives that move one object at a time.
Even such a simple setting is challenging because the space of object configurations is combinatorially large, resulting in long-horizon combinatorial task spaces.
We formulate rearrangement as an offline goal-conditioned reinforcement learning (RL) problem, where the agent is pretrained on a experience buffer of sensorimotor interactions and is evaluated on producing actions for rearranging objects specified in the input image to satisfy constraints depicted in a goal image.

Offline RL methods~\citep{levine2020offline} that do not infer factorized representations of entities struggle to generalize to problems with more objects.
But planning with object-centric methods that do infer entities~\citep{veerapaneni2020entity} is also not easy because the difficulties of long-horizon planning with learned parametric models~\citep{janner2019trust} are exacerbated in combinatorial spaces.

Instead of planning with parametric models, our work takes inspiration from non-parametric planning methods that have shown success in combining neural networks with graph search to generate long-horizon plans.
These methods~\citep{yang2020plan2vec,zhang2018composable,lippi2020latent,emmons2020sparse} explicitly construct a transition graph from the experience buffer and plan by searching through the actions recorded in the graph with a learned distance metric.
The advantage of such approaches is the ability to stitch different path segments from offline data to solve new problems.
The disadvantage is that the non-parametric nature of such methods requires transitions that will be used for solving new problems to have already been recorded in the buffer, making conventional methods, which store entire observations monolithically, ill-suited for combinatorial generalization.
Fig.~\ref{fig:problem}b shows that the same state transition can manifest for different objects and in different contexts, but monolithic non-parametric methods are not constrained to recognize that all scenarios exhibit the same state transition at an abstract level.
This induces an blowup in the number of nodes in the graph.
To overcome this problem, we devise a method that exploits the similarity among state transitions in different contexts.

Our method, \textbf{Neural Constraint Satisfaction} (\methodname)\footnote{See Appdx.~\ref{appdx:name} for an explanation behind this name.},
marries the strengths of non-parametric planning with those of object-centric representations.
Our main contribution is to show that factorizing the traditionally monolithic entity representation into action-invariant features (its \textbf{type}) and action-dependent features (its \textbf{state}) makes it possible during planning and control to reuse action representations for different objects in different contexts, thereby addressing the core combinatorial challenge in object rearrangement.
To implement this factorization,~\methodname constructs a two-level hierarchy (Fig.~\ref{fig:two_level_hierarchy}) to abstract the experience buffer into a graph over state transitions of individual entities, separated from other contextual entities (Fig.~\ref{fig:model}).
To solve new rearrangement problems,~\methodname infers what state transitions can be taken given the current and goal image observations, re-composes sequences of state transitions from the graph, and translates these transitions into actions.

In \S\ref{sec:problem} we introduce a problem formulation that exposes the combinatorial structure of object rearrangement tasks by explicitly modeling the independence, symmetry, and factorization of latent entities.
This reveals two challenges in object rearrangement which we call the \textbf{correspondence problem} and \textbf{combinatorial problem}.
In \S\ref{sec:method} we present~\methodname, a method for controlling an agent that plans over and acts with emergent learned entity representations, as a unified method for tackling both challenges.
We show in \S\ref{sec:experiments} that~\methodname outperforms both state-of-the-art offline RL methods and object-centric shooting-based planning methods in simulated rearrangement problems.

%% file: src/figs/two_level_hierarchy.tex
\begin{wrapfigure}[27]{r}{0.3\textwidth}
  \vspace{-35pt}
  \begin{center}
    \includegraphics[width=0.3\textwidth]{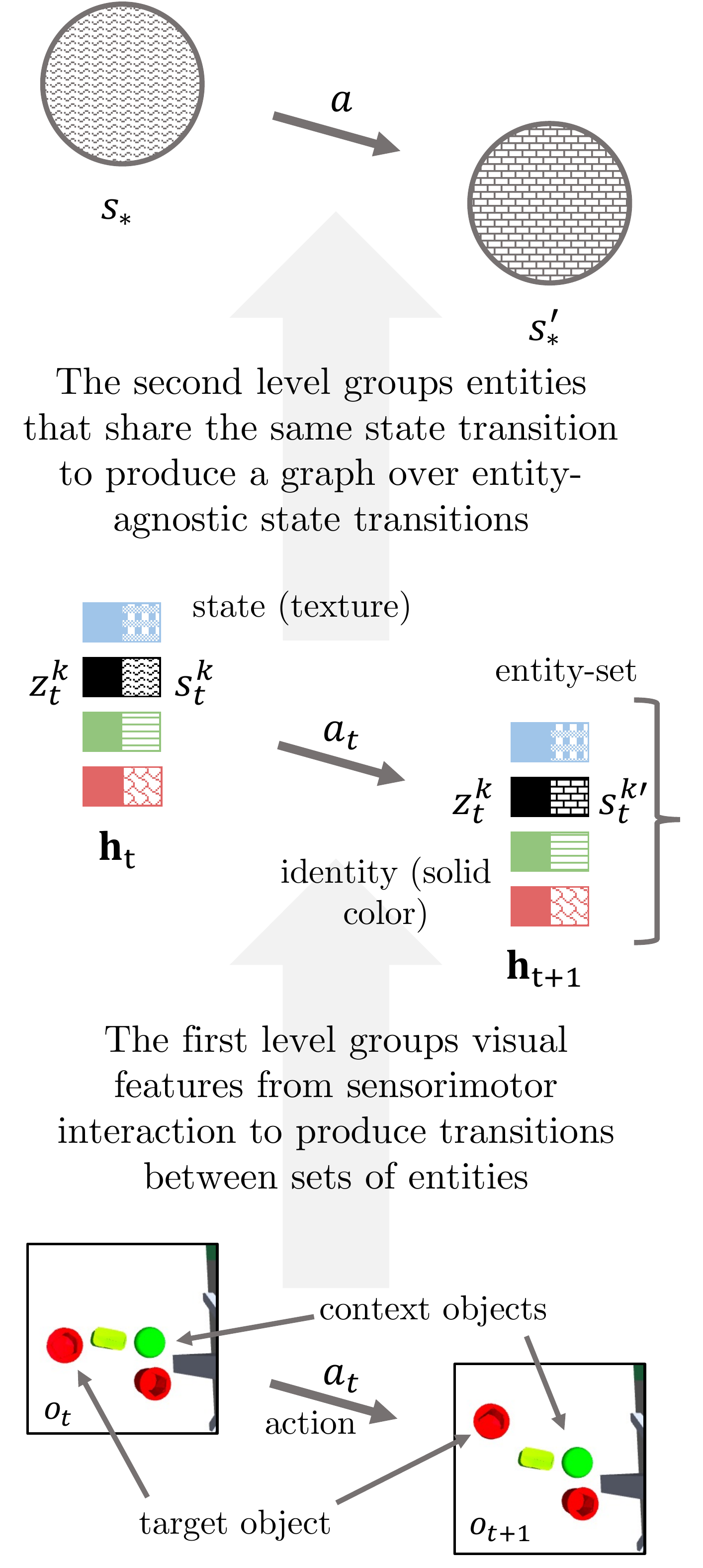}
  \end{center}
  \caption{\small{\methodname uses a two-level hierarchy to abstract sensorimotor interactions into a graph of learned state transitions.
  The affected entity is in black.
  }}
  \label{fig:two_level_hierarchy}
\end{wrapfigure}

%% file: src/related_work.tex
\section{Related Work}
The problem of discovering re-composable representations is generally motivated by combinatorial task spaces.
The traditional approach to enforcing this compositional inductive bias is to compactly represent the task space with MDPs that human-defined abstractions of entities, such as factored MDPs~\citep{Boutilier_exploiting,Boutilier_factored,guestrin2003factored}, relational MDPs~\citep{wang2008relational,guestrin2003generalizing,gardiol_envelope-based_2003}, and object-oriented MDPs~\citep{Diuk_object,AbelHBBOMT15}.
Approaches building off of such symbolic abstractions~\citep{chang2016compositional,battaglia2018relational,zadaianchuk2022self,bapst2019structured,zhang2018composable} do not address the problem of how such entity abstractions arise from raw data.
Our work is one of the first to learn compact representations of combinatorial task spaces directly from raw sensorimotor data.

Recent object-centric methods~\citep{greff2017neural,van2018relational,greff2019multi,greff2020binding,locatello2020object,kipf2021conditional,zoran2021parts,singh2021illiterate} do learn entity representations, as well as their transformations~\citep{alias2021neural,goyal2020object}, from sensorimotor data, but only do so for modeling images and video, rather than for taking actions.
Instead, we study \emph{how well entity-representations can reused for solving tasks}.
\citet{kulkarni2019unsupervised} considers how object representations improve exploration, but we consider the offline setting which requires zero-shot generalization.
\citet{veerapaneni2020entity} also considers control tasks, but their shooting-based planning method suffers from compounding errors as other learned single-step models do~\citep{janner2019trust}, while our hierarchical non-parametric approach enables us to plan for longer horizons.

Non-parametric approaches have recently become popular for long horizon planning~\citep{yang2020plan2vec,zhang2018composable,lippi2020latent,emmons2020sparse,zhang2021worldmodel}, but the drawback of these approaches is they represent the entire scenes monolithically, which causes a blowup of nodes in combinatorial task spaces, making it infeasible for these methods to be applied in rearrangement tasks that require generalizing to novel object configurations with different numbers of objects.
Similar to our work, \citet{huang2019neural} also tackles rearrangement problems by searching over a constructed latent task graph, but they require a demonstration during deployment time, whereas~\methodname does not because it reuses context-agnostic state transitions that were constructed during training.
\cite{zhang2021worldmodel} conducts non-parametric planning directly on abstract subgoals rather than object-centric states --- while similar, the downside of using subgoals rather than abstract states is that those subgoals are not used to represent equivalent states and therefore cannot generalize to new states at test time. Our method, NCS, captures both reachability between known states and new, unseen states that can be mapped to the same abstract state.

%% file: src/problem.tex
\section{Goal-Conditioned Reinforcement Learning with Entities} \label{sec:problem}
This section introduces a set of modifications to the standard goal-conditioned partially observed Markov decision process (POMDP) problem formulation that explicitly expose the combinatorial structure of object rearrangement tasks of the following kind: ``Sequentially move a subset (or all) of the objects depicted in the current observation $o_1$ to satisfy the constraints depicted in the goal image $o_g$.''
We assume an offline RL setting, where the agent is trained on a buffer of transitions $\{(o_1, a_1, ... a_{T-1}, o_T)\}_{n=1}^N$ and evaluated on tasks specified as $(o_1, o_g)$.

The standard POMDP problem formulation assumes an observation space $\mathcal{O}$, action space $\mathcal{A}$, latent space $\mathcal{H}$, goal space $\mathcal{G}$, observation function  $E: \mathcal{H} \rightarrow \mathcal{O}$, transition function $P: \mathcal{H} \times \mathcal{A} \rightarrow \mathcal{H}$, and reward function  $R: \mathcal{H} \times \mathcal{G} \rightarrow \mathbb{R}$.
Monolithically modeling the latent space this way does not expose commonalities among different scenes, such as scenes that contain objects in the same location or scenes with multiple instances of the same type of object.
This prevents us from designing control algorithms that exploit these commonalities to collapse the combinatorial task space.

\input{src/figs/problem}
To overcome this issue, we introduce structural assumptions of independence, symmetry, and factorization to the standard formulation. 
The \emph{independence} assumption encodes the intuitive property that objects can be acted upon without affecting other objects.
This is implemented by decomposing the latent space into independent subspaces as $\mathcal{H} = \mathcal{H}^1 \times ... \times \mathcal{H}^K$, one for each independent degree of freedom (e.g. object) in the scene.
The \emph{symmetry} assumption encodes the property that the the same physical laws apply to all objects.
This is implemented by constraining the observation, transition, and reward functions to be shared across all subspaces, thereby treating $\mathcal{H}^1 = ... = \mathcal{H}^K$.
We define an \textbf{entity}\footnote{We use ``object'' to refer to an independent degree of freedom in the environment, and ``entity'' to refer to the agent's representation of the object.} $h \in \mathcal{H}^k$ as a member of such a subspace, and an \textbf{entity-set} as the set of entities $\mathbf{h} = (h^1, ..., h^K)$ that explain an observation, similar to~\citet{Diuk_object,weld_solving_nodate}.
Lastly, the \emph{factorization} assumption encodes that each subspace can be decomposed as $\mathcal{H}^k = \mathcal{Z} \times \mathcal{S}$, where $z \in \mathcal{Z}$ represents an entity's action-invariant features like appearance, and $s \in \mathcal{S}$ represents its action-dependent features like location.
We call $z$ the \textbf{type} and $s$ the \textbf{state}.

Introducing these assumptions solves the problem of modeling the commonalities among different scenes stated above.
It allows us to describe scenes that contain objects in the same location by assigning entities in different scenes to share the same state $s$.
It allows us to describe a scene with multiple instances of the same type of object by assigning multiple entities in the scene to share the same type $z$.
This formulation also makes it natural to express goals as a set of constraints $\mathbf{h}_g = (h^1_g, ..., h^k_g)$. 
To solve a task is to take actions that transform the subset of entities in the initial observation $o_1$ whose types are given by $\mathbf{z}_g$ to new states specified by $\mathbf{s}_g$.

Exposing this structure in our problem formulation gives us a language for designing methods that represent entities in an independent, symmetric, and factorized way and that use these three properties to collapse the combinatorial task space.
These methods need to solve two problems: the \textbf{correspondence problem} of learning to represent entities in this way and the \textbf{combinatorial problem} of using these properties to make planning tractable.
The correspondence problem is hard because it assumes no human supervision of what the entities are.
It also goes beyond problems solved by existing object-centric methods for images and videos
because it involves action: it requires representing entities such that there is a correspondence between how the agent models how its actions affect entities and how its actions actually affect objects in the environment.
The combinatorial problem goes beyond problems solved by methods for solving 
object-oriented MDPs,
relational MDPs,
and factorized MDPs
because it requires the agent to recognize whether and how previously observed state transitions can be used for new problems, using learned, rather than human-defined, entity representations.
The natural evaluation criterion for both problems is to test to what extent an agent can zero-shot-generalize to solve rearrangement tasks involving new sets of object configurations that aree disjoint from the configurations observed in training, assuming that the training configurations have collectively covered $\mathcal{Z}$ and $\mathcal{S}$.
Our experiments in \S\ref{sec:experiments} test exactly this. 

\paragraph{Simplifying assumptions}
To focus on the combinatorial nature of rearrangement, we are not interested in low-level manipulation, so we represent each action as $(w, \Delta w)$, where $w$ are  Cartesian coordinates $w = (x, y, z)$.
We assume actions sparsely affect one entity at a time and how an action affects an object's state does not depend on its identity.
We are not interested in handling occlusion, so we assume that objects are constrained to the $xy$ plane or $xz$ plane and are directly visible to the camera.
Following prior work~\citep{hansen2022bisimulation,castro2009equivalence}, we make a \emph{bisimulation} assumption that the state space can be partitioned into a finite set of equivalence classes, and that there is one action primitive that transitions between each pair of equivalence classes.
Lastly, we assume objects can be moved independently.
Preliminary experiments suggest that~\methodname can be augmented to support tasks like block-stacking that involve dependencies among objects, but how to handle these dependencies would warrant a standalone treatment in future work.

%% file: src/figs/problem.tex
\begin{figure}
    \centering
    \vspace{-20pt}
    \includegraphics[width=\textwidth]{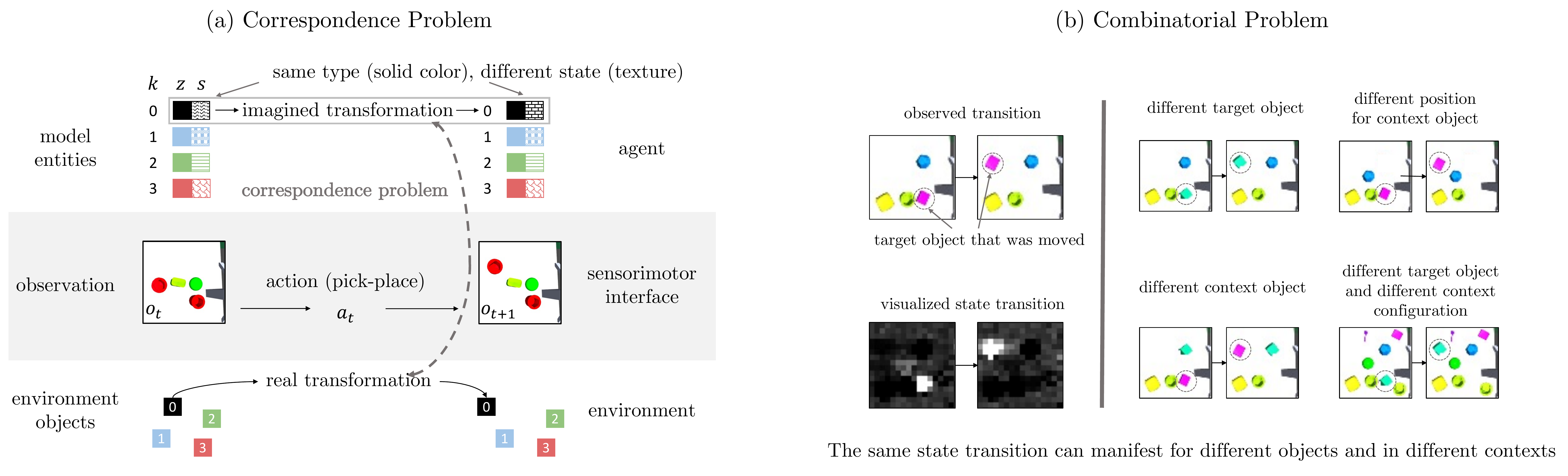}
    \caption{\small{\textbf{Solving object rearrangement requires solving two challenges}. 
    (a) The \textbf{correspondence problem} is the problem of abstracting raw sensorimotor signal into representations of entities such that there is a correspondence between how an agent intervenes on an entity and how its action affects an object in the environment.
    $k$ denotes the index of the entity, $z$ denotes its type (shown with solid colors), and $s$ denotes its state (shown with textures).
    The entity representing the moved object is in black.
    (b) The \textbf{combinatorial problem} is the problem of representing the combinatorial task space in a way that enables an agent to transfer knowledge of a given state transition (indicated by the dotted circle) to different contexts.
    }}
    \label{fig:problem}
    \vspace{-15pt}
\end{figure}

%% file: src/method.tex
\input{src/figs/model}

\section{Neural Constraint Satisfaction} \label{sec:method}
In \S\ref{sec:problem} we introduced a structured problem formulation for object rearrangement and reduced it to solving the correspondence and combinatorial problems.
We now present our method, Neural Constraint Satisfaction (\methodname) as a method for controlling an agent that plans over and acts with a state transition graph constructed from learned entity representations.
This section is divided into two parts: modeling and control.
The modeling part is further divided into two parts: representation learning and graph construction.
The representation learning part addresses the correspondence problem, while the graph construction and control parts address the combinatorial problem.

\subsection{Modeling}
The modeling component of~\methodname abstracts the experience buffer into a factorized state transition graph that can be reused across different rearrangement problems.
Below we describe how we first train an object-centric world model to infer entities that are independent, symmetric, and factorized and then construct the state transition graph by clustering entities with similar state transitions.
These two steps comprise a two-level abstraction hierarchy over the raw sensorimotor transitions.

\paragraph{Level 1: representation learning}
The first level concerns the unsupervised learning of entity representations that factorizes into their action-invariant features (their \textbf{type}) and their action-dependent features (their \textbf{state}).
Concretely our goal is to model a video transition $o_t, a_t \rightarrow o_{t+1}$ as a transition over entity-sets $\mathbf{h}_t, a_t \rightarrow \mathbf{h}_{t+1}$, where each entity $h^k$ is factorized as a pair $h^k = (z^k, s^k)$.
Given our setting where an action moves only a single object in the environment at a time, successful representation learning implies three criteria: (1) the world model properly identifies the individual entity $h^k$ corresponding to the moved object, (2) only the state $s^k$ of that entity should change, while its type $z^k$ should remain unaffected, and (3) other entity representations $h^{\neq k}$ should also remain unaffected.
Criteria (1) and (3) rule out standard approaches that represent an entire scene with a monolithic representation, so we need an object-centric world model instead of a monolithic world model.
But criterion (2) rules out standard object-centric world models (e.g.~\citep{veerapaneni2020entity,elsayed2022savi++,singh2022simple}), which do not decompose entity representations into action-invariant and action-dependent features.

Because the parameters of a mixture model are independent and symmetric by construction, we propose to construct our factorized object-centric world model as an equivariant sequential Bayesian filter with a mixture model as the latent state, where entity representations are the parameters of the mixture components.
Recall that a filter consists of two major components, latent estimation and latent prediction.
We implement latent estimation with the state-of-the-art slot attention (SA)~\citep{locatello2020slot}, based on the connection between mixture components and SA slots~\citep{chang2022object}.
We implement latent prediction with the transformer decoder (TFD) architecture~\citep{vaswani2017attention} because TFD is equivariant with respect to its inputs.
We denote the SA \texttt{slot}s as $\bm{\lambda}$ and SA \texttt{attn} masks as $\bm{\alpha}$.
We split each \texttt{slot} $\lambda \in \mathbb{R}^n$ into two halves $\lambda^z \in \mathbb{R}^{\frac{n}{2}}$ and $\lambda^s \in \mathbb{R}^{\frac{n}{2}}$.
Given observations $o$ and actions $a$, we embed the actions as $\tilde{a}$ with an feedforward network and implement the filter as:
\begin{align*}
    \hat{\bm{\lambda}}_1 &\sim \text{Gaussian} \qquad
    &\hat{\bm{\lambda}^{s}}_{t+1} &= TFD\left(
        \text{queries}=\bm{\lambda}^{s}_{t}, 
        \text{keys/values}=\left[\bm{\lambda}^{s}, \tilde{a}_t\right]
        \right) \\
    \bm{\lambda}_{t}, \bm{\alpha}_{t} &= SA\left(\hat{\bm{\lambda}}_{t}, o_{t}\right) \qquad
    &\hat{\bm{\lambda}}_{t+1} &= \left[\bm{\lambda}^{z}_t, \hat{\bm{\lambda}^{s}}_{t+1}\right]
\end{align*}
where $[ \cdot, \cdot ]$ is the concatenation operator, $\hat{\bm{\lambda}}$ is the output of the latent prediction step, and $\bm{\lambda}$ is the output of the latent estimation step.
We embed this filter inside the SLATE backbone~\citep{singh2022illiterate} and call this implementation \textbf{dynamic SLATE} (dSLATE).
For a background on SLATE, as well as dSLATE hyperparamters, see Appdx.~\ref{appdx:slate_background}.

By constructing $\hat{\bm{\lambda}^{z}}_{t+1}$ as a copy of $\bm{\lambda}^{z}_t$, dSLATE enforces the information contained $\bm{\lambda}^{z}$ to be action-invariant, hence we treat $\bm{\lambda}^{z}$ as dSLATE's representation of the entities' types.
As for the entities' states, either the action-dependent part of the slots $\bm{\lambda}^{s}$ or the attention masks $\alpha$ can be used.
Using $\alpha$ may be sufficient and more intuitive to analyze if all objects looks similar and there is no occlusion, while $\bm{\lambda}^{s}$ may be more suitable in other cases, and we provide an example of each in the experiments.
To simplify notation going forward and connect with the notation in \S\ref{sec:problem}, we use $\mathbf{h}$ to refer to $(\bm{\lambda}, \bm{\alpha})$, use $\bm{z}$ to refer to $\bm{\lambda}^{z}$, and use $\bm{s}$ to refer to $\bm{\lambda}^{s}$ or $\bm{\alpha}$.
Thus by construction dSLATE satisfies criterion (2).
Empirically we observe that it satisfies criterion (1) as well as SLATE does, and that TFD learns to sparsely edit $\bm{\lambda}^{s}_t$, thereby satisfying criterion (3).

\paragraph{Level 2: graph construction} \label{sec:build_graph}
Having produced from the first level a buffer of entity-set transitions ${\{\mathbf{h}_t, a_t \rightarrow \mathbf{h}_{t+1}\}_{n=1}^N}$, the goal of the second level (Fig.~\ref{fig:model}b) is to use this buffer to construct a factorized state transition graph.
The key to solving the combinatorial problem is to construct the edges of this graph to represent not state transitions of entire entity-sets (i.e. ${\mathbf{s}_t, a_t \rightarrow \mathbf{s}_{t+1}}$) as prior work does~\citep{zhang2018composable}, but state transitions of \emph{individual entities} (i.e. ${s^k_t, a_t \rightarrow s^k_{t+1}}$).
Constructing edges over transitions for individual entities rather than entity sets enables the same transition to be reused with different context entities present.
Constructing edges over state transitions instead of entity transitions enables the same transition to be reused across entities with different types.
This would enable the agent to recompose sequences of previously encountered state transitions for solving new rearrangement problems with different entities in different contexts.
Henceforth our use of ``state'' refers specifically to the state of individual entity unless otherwise stated. 

Given our bisimulation assumption that states can be partitioned into a finite number of groups, we construct our graph such that nodes represent equivalence classes among individual states and the edges represent actions that transform a state from one equivalence class to another.
To implement this we cluster state transitions of individual entities in the buffer, which reduces to clustering the states of individual entities before and after the transition.
We treat each cluster centroid as a node in the graph, and an edge between nodes is tagged with the single action that transforms one node's state to another's.
The algorithm for constructing the graph is shown in Alg.~\ref{alg:build_graph} and involves three steps: (1) isolating the state transition of an individual entity from the state transition of the entity-set, (2) creating graph nodes from state clusters, and (3) tagging graph edges with actions.

\input{src/figs/building_the_graph}

The first step is to identify which object was moved in each transition, i.e. identifying the entity $h^k$ that dSLATE predicted was affected by $a_t$ in the transition $(\mathbf{h}_t, a_t, \mathbf{h}_{t+1})$.
We implement a function \texttt{isolate} that achieves this by solving $k = \argmax_{k' \in \{1, ..., K\}} d(s^{k'}_t, s^{k'}_{t+1})$ to identify the index of the entity whose state has most changed during the transition, where $d(\cdot, \cdot)$ is a distance function, detailed in Table~\ref{tab:FACTS_hyperparameters} of the Appendix.
This converts the buffer of transitions over entity-sets ${\mathbf{h}_t, a_t \rightarrow \mathbf{h}_{t+1}}$ into a buffer of transitions over individual entities ${h^k_t, a_t \rightarrow h^k_{t+1}}$.

The second step is to cluster the states before and after each transition.
We implement a function \texttt{cluster} that uses K-means to returns graph nodes as the centroids $\{s_*\}_{m=1}^M$ of these state clusters.

The third step is to connect the nodes with edges that record actions that actually were taken in the buffer to transform one state to the next.
We implement a function \texttt{bind} that, given entity $h^k$, returns the index $[i]$ of the centroid $s_*$ that is the nearest neighbor to the entity's state $s^k$.
For each entity transition $(h_t^k, a_t, h_{t+1}^k)$ we \texttt{bind} entity $h_t^k$ and $h_{t+1}^k$ to their associated nodes $s_*^{[i]}$ and $s_*^{[j]}$ and create an edge between $s_*^{[i]}$ and $s_*^{[j]}$ tagged with action $a$, overwriting previous edges based on the assumption that with a proper clustering there should only be one action per pair of nodes.

In our experiments both \texttt{cluster} and \texttt{bind} use the same distance metric (see Table~\ref{tab:dslate_hyperparameters} in the Appendix), but other clustering algorithms and distance metrics can also be used.
Our experiments (Fig.~\ref{fig:robogym_analysis}) also show that it is also possible to have more than one action primitive per pair of nodes as long as these actions all map between states bound to the same pair of nodes.

\input{src/figs/planning_and_control}

\subsection{Control}
To solve new rearrangement problems, we re-compose sequences of state transitions from the graph.
Specifically, the agent decomposes the rearrangement problem into a set of per-entity subproblems (e.g. initial and goal positions for individual objects), searches the transition graph for a transition that transforms the current entity's state to its goal state, and executes the action tagged with this transition in the environment.
This problem decomposition is possible because the transitions in our graph are constructed to be agnostic to type and context, enabling different rearrangement problems to share solutions to the same subproblems.
The core challenge in deciding which transitions to compose is in determining which transitions are \emph{possible} to compose.
That is, the agent must determine which nodes in the graph correspond to the given goal constraints and which nodes correspond to the entities in the current observation, but the current entities $\mathbf{h}_t$  and goal constraints $\mathbf{h}_g$ must themselves be inferred from the current and goal observations $o_t$ and $o_g$, requiring the agent to infer both what to do and how to do it purely from its sensorimotor interface.

\input{src/figs/action_selection}

Our approach takes four steps, summarized in Alg.~\ref{alg:use_graph} and Fig.~\ref{fig:planning_and_control}, with further details in Appdx.~\ref{appdx:action_selection}.
In the first step, we use dSLATE to infer $\mathbf{h}_t$ and $\mathbf{h}_g$ from $o_t$ and $o_g$ (e.g. the positions and types of all objects in the initial and goal images).
In the second step (Fig.~\ref{fig:planning_and_control}b), because of the permutation symmetry among entities, we find a bipartite matching that matches each entities in $h_g^j$ with a corresponding entity in $h_t^k$ that shares the same type and permute the indices $k$ of $\mathbf{h}_t$ to match those of $\mathbf{h}_g$.
We implement a function \texttt{align} that uses the Hungarian algorithm to perform this matching over $(z^1_t, ... z^K_t)$ and $(z^1_g, ... z^K_g)$, with Euclidean distance as the matching cost.
The third step selects which goal constraint $h^k_g$ to satisfy next (Fig.~\ref{fig:planning_and_control}c).
W implement this \texttt{select-constraint} procedure by determining which constraint $h^k_g$ has the highest difference in state with its counterpart $h^k_t$, which reduces to solving the same argmax problem as in \texttt{isolate} with the same distance function used in \texttt{isolate}.
The last step chooses an action given the chosen goal constraint $h^k_g$ and its counterpart  $h^k_t$, by \texttt{bind}ing $h^k_t$ and $h^k_g$ to the graph based on their state components and returning the action tagged to the edge between their respective nodes (Fig.~\ref{fig:planning_and_control}d).
If an edge does not exist between the inferred nodes, then we simply take a random action.

%% file: src/figs/model.tex
\begin{figure}
    \centering
    \vspace{-20pt}
    \includegraphics[width=\textwidth]{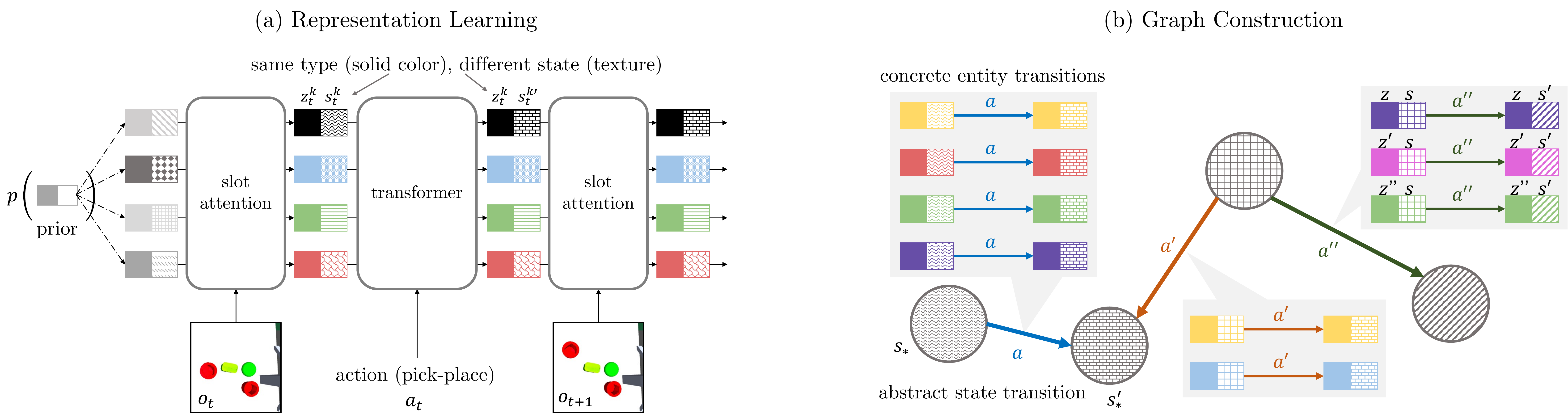}
    \caption{\small{\textbf{Modeling}.
    \methodname constructs a two-level abstraction hierarchy to model transitions in the experience buffer.
    (a) \textbf{Level 1:}~\methodname learns to infer a set of entities from sensorimotor transitions with pick-and-move actions, in which one entity is moved per transition.
    We enforce that the type $z$ (shown with solid colors) of an entity remains unchanged between time-steps.
    The GPT dynamics model learns to sparsely predict the states $s$ (shown with textures) of the entities at the next time-step.
    \emph{This addresses the correspondence problem by forcing the network to use predict and reconstruct observations through the entity bottleneck.}
    (b) \textbf{Level 2:}~\methodname abstracts transitions over entity-sets into transitions over states of individual entities, constructing a graph where states are nodes and transitions between them are edges.
    This is done by clustering entity transitions that share similar initial states and final states.
    \emph{This addresses the combinatorial problem by making it possible for state transitions to reused for different entity types and with different context entities.
    }
    }}
    \label{fig:model}
    \vspace{-10pt}
\end{figure}

%% file: src/figs/building_the_graph.tex
\begin{wrapfigure}[20]{r}{0.6\linewidth}
\vspace{-20pt}
\begin{minipage}{0.6\textwidth}
\begin{algorithm}[H]
  \caption{Building the Graph}\label{alg:build_graph}
\small
  \begin{algorithmic}[1]
    \State \textbf{input} \texttt{model}, \texttt{buffer}
    \For{$\{(o_t, a_t, o_{t+1})\}_n$ in \texttt{buffer}} 
        \State{\textcolor{gray}{\footnotesize{\# infer entities from transition}}}
        \State $\{(\mathbf{h}_t, a_t, \mathbf{h}_{t+1})\}_n \gets \texttt{model}\left(\{o_t, a_t, o_{t+1}\}_n\right)$.
        \State{\textcolor{gray}{\footnotesize{\# identify which entity changed in transition}}}
        \State $\{(h^k_t, a_t, h^k_{t+1})\}_n\gets \texttt{isolate}\left(
        \{(\mathbf{h}_t, a_t, \mathbf{h}_{t+1})\}_n\right)$
    \EndFor
    \State{\textcolor{gray}{\footnotesize{\# partition transitions by clustering entities}}}
    \State $\{s_*\}_{m=1}^M \gets \texttt{cluster}\left(\{(s^k_t, a_t, s^k_{t+1})\}_{n=1}^N\right)$ 
    \State{\textcolor{gray}{\footnotesize{\# transitions between clusters are edges}}}
    \State \textbf{initialize} \texttt{graph} with nodes $s_*^{[m]}$, for $m \in [1:M]$
    \ForEach{$\{(h^k_t, a_t, h^k_{t+1})\}_n$}
        \State{\textcolor{gray}{\footnotesize{\# infer cluster assignments}}}
        \State $[i], [j]  \gets \texttt{bind}\left(h_{t}^k\right), \texttt{bind}\left(h_{t+1}^k\right)$
        \State{\textcolor{gray}{\footnotesize{\# tag edge with action $a_t$}}}
        \State \texttt{graph.edges}$[i, j] \gets \texttt{create-edge}\left(s_*^{[i]} \overset{a_t}{\rightarrow} s_*^{[j]}\right)$
    \EndFor
    \State \Return \texttt{graph}
  \end{algorithmic}
\end{algorithm}
\end{minipage}
\vspace{-10pt}
\end{wrapfigure}

%% file: src/figs/planning_and_control.tex
\begin{figure}
    \centering
    \includegraphics[width=0.9\textwidth]{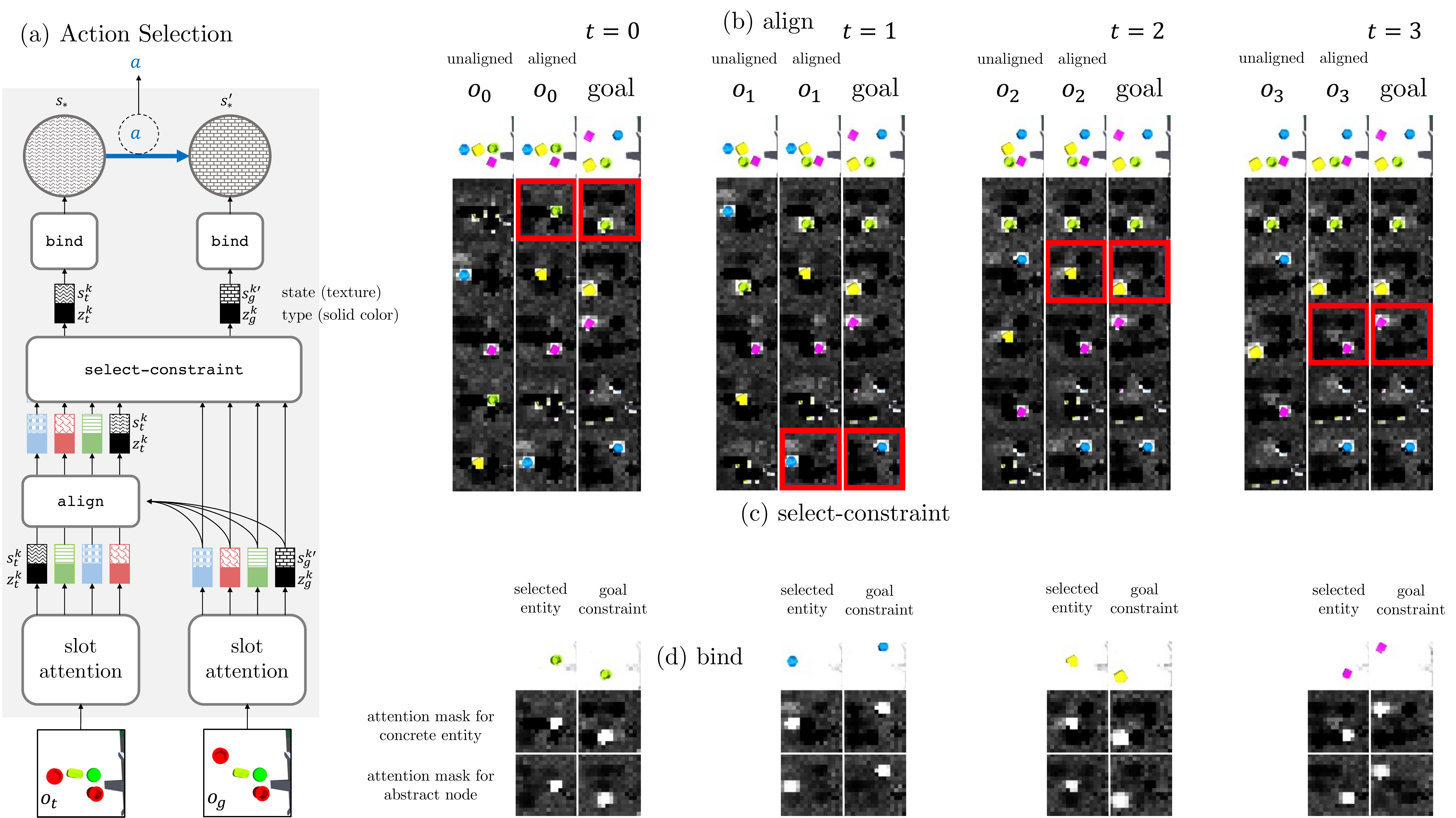}
    \caption{\small{\textbf{Planning and control}.
    Given a rearrangement problem specified only by the current and goal observations $(o_0, o_g)$,~\methodname decomposes the rearrangement problem into one subproblem $(o_t, o_g)$ per entity.
    (a) shows the computations~\methodname uses to solve each subproblem and (b-d) show these steps in context.
    For each subproblem $(o_t, o_g)$,~\methodname infers entities from both the current and goal observations.
    The states of the goal entities indicate constraints on the desired locations of the current entities.
    (b) ~\methodname aligns the indices of the current entities to those of the goal entities with corresponding types.
    (c) It selects the index $k$ of the next goal constraint $s_g^k$ to satisfy, as indicated by the red box.
    The selected goal constraint and current entity are also colored black in (a), and note that their types are the same but states are different: we want to choose the action to transform the state of the current entity to the state of the goal constraint.
    (d) It binds the selected goal constraint and its corresponding current entity to nodes $s_*$ and $s_*'$ in the transition graph.
    Lastly, it identifies the edge connecting those two nodes and executes the action tagged to that edge in the environment.
    }}
    \label{fig:planning_and_control}
    \vspace{-10pt}
\end{figure}

%% file: src/figs/action_selection.tex
\begin{wrapfigure}[17]{r}{0.45\linewidth}
\vspace{-13pt}
\begin{minipage}{0.45\textwidth}
\begin{algorithm}[H]
  \caption{Action Selection}\label{alg:use_graph}
  \begin{algorithmic}[1]
    \State \textbf{given} \texttt{model}, \texttt{graph}
    \State \textbf{input} goal $o_g$, observation $o_t$
    \State{\textcolor{gray}{\footnotesize{\# infer goal constraints and current entities}}}
    \State $\mathbf{h_g}, \mathbf{h_t} \leftarrow \texttt{model}\left(o_g\right), \texttt{model}\left(o_t\right)$
    \State{\textcolor{gray}{\footnotesize{align entity indices of $\mathbf{h_t}$ with those of $\mathbf{h_g}$}}}
    \State $\pi \gets \texttt{align}\left(\mathbf{h_t}, \mathbf{h_g}\right)$
    \State{\textcolor{gray}{\footnotesize{permute indices of $\mathbf{h_t}$ according to $\pi$}}}
    \State $\mathbf{h_t} \gets (h_t^{\pi[1]}, ..., h_t^{\pi[K]})$
    \State{\textcolor{gray}{\footnotesize{identify $k$th goal constraint to satisfy next}}}
    \State $k \leftarrow \texttt{select-constraint}\left(\mathbf{h_t}, \mathbf{h_g}\right)$
    \State{\textcolor{gray}{\footnotesize{infer cluster assignments}}}
    \State $[i], [j]  \gets \texttt{bind}\left(h_{t}^k\right), \texttt{bind}\left(h_{g}^k\right)$
    \State{\textcolor{gray}{\footnotesize{action that transforms node $[i]$ to node $[j]$}}}
    \State \Return \texttt{graph.edges}$[i, j]$\texttt{.action}
  \end{algorithmic}
\end{algorithm}
\end{minipage}
\end{wrapfigure}

%% file: src/experiments.tex
\section{Experiments} \label{sec:experiments}
We have proposed~\methodname as a solution to the object rearrangement problem that addresses two challenges:~\methodname addresses the correspondence problem by learning a factorized object-centric world model with dSLATE and it addresses the combinatorial problem by abstracting entity representations into a queryable state transition graph.
Now we test~\methodname's efficacy in solving both problems.

The key question is whether~\methodname is better than state-of-the-art offline RL algorithms in generalizing over combinatorially-structured task spaces from perceptual input.
As stated in \S\ref{sec:problem}, the crucial test for answering this question is to evaluate all methods on solving new rearrangement problems with a disjoint set of object configurations from those encountered during training.
The most straightforward way to find a disjoint subset of the combinatorial space is to evaluate with a novel number of objects.
We compare~\methodname to several offline RL baselines and ablations on two rearrangement environments and find a significant gap in performance between our method and the next best method.

\input{src/figs/experiment_tasks}

\textbf{Environments.}
In \emph{block-rearrange} (Fig.~\ref{fig:environments}a), all objects are the same size, shape, and orientation.
$\mathcal{S}$ covers 16 locations in a grid. 
$\mathcal{Z}$ is the continuous space of red-green-blue values from $0$ to $1$.
\emph{robogym-rearrange} (Fig.~\ref{fig:environments}b) is adapted from the~\citet{robogym2020} rearrange environment and removes the assumptions from \emph{block-rearrange} that all objects have the same size, shape, and orientation.
The objects are uniformly sampled from a set of 94 meshes consisting of the YCB object set \citep{calli2015ycb} and a set of basic geometric shapes, with colors sampled from a set of 13.
Although locations are not pre-defined in \emph{robogym-rearrange} as in \emph{block-rearrange}, in practice there is a limit to the number of ways to arrange objects on the table to still be visible to the camera, which makes the bisimulation still a reasonable assumption here.
For \emph{block-rearrange} we use the SA attention mask $\alpha$ as the state $s$, and for \emph{robogym-rearrange} we use the action-dependent part of the SA slot $\bm{\lambda}^s$ as the state $s$.
Further environment details are in Appdx.~\ref{appdx:environment_details}.

\textbf{Experimental setup.}
We evaluate two settings: \emph{complete} and \emph{partial}.
In the \emph{complete} setting, the goal image shows all objects in new locations.
The \emph{partial} setting is underspecified: only a subset of objects have associated goal constraints (Fig.~\ref{fig:environments}b).
In \emph{block-rearrange}, all constraints are unsatisfied in the start state.
In \emph{robogym-rearrange}, four constraints are unsatisfied in the start state.
Our metric is the \emph{fractional success rate}, the average change in the number of satisfied constraints divided by the number of initially unsatisfied constraints.

The experiences buffer consists of 5000 trajectories showing 4 objects.
We evaluate on 4-7 objects for 100 episodes across 10 seeds.
Even if we assume full access to the underlying state space, the task spaces are enormous: with $|S|$ object locations and $k$ objects, the number of possible trajectories over object configurations of $t$ timesteps is ${|S| \choose k} \times (k \times (|S| - k))^t$, which amounts to searching over more than $10^{16}$ possible trajectories 
for the complete specification setting of \emph{block-rearrange} with $k=7$ objects (see Appdx.~\ref{appdx:combinatorial_space} for derivation).
Our setting of assuming access to only pixels makes the problem even harder.

\textbf{Baselines.}
The claim of this paper are that, for object rearrangement, (1) object-centric methods fare better than monolithically-structured offline RL methods (2) non-parametric graph search fares better than parametric planning for object rearrangement and (3) a factorized graph search over state transitions of individual entities fares better than a non-factorized graph search over state transitions over entire entity-sets.
To test (1), we compare with state-of-the-art pixel-based behavior cloning (BC) and implicit Q-learning (IQL) implementations based off of~\cite{jaxrl}.
To test (2), we compare against a version of object-centric model predictive control (MPC)~\citep{veerapaneni2020entity} that uses the cross entropy method over dSLATE rollouts.
To test (3), we compare against an ablation (abbrv. NF, for ``non-factorized'') that constructs a graph with state transitions of entity-sets than of individual states.
Our last baseline just takes random actions (Rand).
Baseline implementation details are in Appdx.~\ref{sec:baseline_implementation_details}.

\subsection{Results}
Figure~\ref{tab:quantiative_results} shows that~\methodname performs significantly better than all baselines (about a 5-10x improvement), thereby refuting the alternatives to our claims in our experimental context.
Most of the baselines perform no better or only slightly better than random.
We observe that it is indeed difficult to perform shooting-based planning with an entity-centric world model trained to predict a single step forward~\citep{janner2019trust}: the MPC baseline performs poorly because its rollouts are poor, and it is significantly more computationally expensive to run (11 hours instead of 20 minutes).
We also observe that the NF ablation performs poorly, showing the importance of factorizing the non-parametric graph search.
Additional results are in Appdx.~\ref{appdx:additional_results}, with limitations in Appdx.~\ref{appdx:limitations}

\begin{table}[!htb]
    \centering
    \small
    \caption{
    This table compares~\methodname with various baselines in the complete and partial evaluation settings of \emph{block-rearrange} and \emph{robogym-rearrange}.
    The methods were trained on 4 objects and evaluated on generalizing to 4, 5, 6, and 7 objects.
    We report the fractional success rate, with a standard error computed over 10 seeds.
    }
    \label{tab:quantiative_results}
    \begin{subtable}{.48\linewidth}
    \raggedleft
      \vspace{-5pt}
      \caption{
      \label{tab:block_rearrange_combinatorial_generalization}
      \emph{block-rearrange}, complete specification.}
      \vspace{-5pt}
        \resizebox{\textwidth}{!}{\begin{tabular}{ c c c c c } 
 Method & 4 & 5 & 6 & 7 \\
 \toprule
 \textbf{\methodname (ours)} & \textbf{0.94} \tiny{$\pm$ 0.01} & \textbf{0.93} \tiny{$\pm$ 0.00} & \textbf{0.93} \tiny{$\pm$ 0.00} & \textbf{0.89} \tiny{$\pm$ 0.00} \\
 Rand  & 0.06 \tiny{$\pm$ 0.02} & 0.07 \tiny{$\pm$ 0.03} & 0.07 \tiny{$\pm$ 0.03} & 0.08 \tiny{$\pm$ 0.03} \\
 MPC  & 0.16 \tiny{$\pm$ 0.06} & 0.12 \tiny{$\pm$ 0.04} & 0.11 \tiny{$\pm$ 0.04} & 0.10 \tiny{$\pm$ 0.03} \\
 NF  & 0.07 \tiny{$\pm$ 0.03} & 0.06 \tiny{$\pm$ 0.02} & 0.07 \tiny{$\pm$ 0.02} & 0.08 \tiny{$\pm$ 0.03} \\
 IQL   & 0.07 \tiny{$\pm$ 0.01}  &  0.03 \tiny{$\pm$ 0.00}  &  0.02 \tiny{$\pm$ 0.00}  &  0.02 \tiny{$\pm$ 0.00} \\ 
 BC  &  0.03 \tiny{$\pm$ 0.00}  &  0.02 \tiny{$\pm$ 0.00}  &  0.01 \tiny{$\pm$ 0.00}  &  0.01 \tiny{$\pm$ 0.00}  \\ 
\end{tabular}}
    \end{subtable}
    \quad
    \begin{subtable}{.48\linewidth}
      \raggedright
        \small
      \vspace{-5pt}
        \caption{\label{tab:block_stacking_combinatorial_generalization} \emph{block-rearrange}, complete specification.}
        \vspace{-5pt}
        \resizebox{\textwidth}{!}{\begin{tabular}{ c c c c c } 
 Method  & 4 & 5 & 6 & 7 \\
 \toprule
\textbf{\methodname (ours)} & \textbf{0.89} \tiny{$\pm$ 0.01} & \textbf{0.86} \tiny{$\pm$ 0.01} & \textbf{0.78} \tiny{$\pm$ 0.01} & \textbf{0.70} \tiny{$\pm$ 0.01} \\ 
Rand & 0.06 \tiny{$\pm$ 0.02} & 0.08 \tiny{$\pm$ 0.03} & 0.08 \tiny{$\pm$ 0.03} & 0.08 \tiny{$\pm$ 0.03} \\
MPC & 0.13 \tiny{$\pm$ 0.05} & 0.11 \tiny{$\pm$ 0.04} & 0.10 \tiny{$\pm$ 0.04} & 0.08 \tiny{$\pm$ 0.03} \\
NF & 0.06 \tiny{$\pm$ 0.03} & 0.07 \tiny{$\pm$ 0.03} & 0.08 \tiny{$\pm$ 0.03} & 0.07 \tiny{$\pm$ 0.03} \\
IQL & 0.01 \tiny{$\pm$ 0.01} & 0.07 \tiny{$\pm$ 0.01} & 0.05 \tiny{$\pm$ 0.01} & 0.05 \tiny{$\pm$ 0.00} \\ 
BC & 0.05 \tiny{$\pm$ 0.01} & 0.04 \tiny{$\pm$ 0.00} & 0.03 \tiny{$\pm$ 0.00} & 0.03 \tiny{$\pm$ 0.00} \\ 
\end{tabular}}
    \end{subtable}
\begin{subtable}{.48\linewidth}
    \raggedleft
     \vspace{5pt}
      \caption{\label{tab:robogym_rearrange_combinatorial_generalization}\emph{robogym-rearrange}, complete specification.}
      \vspace{-5pt}
        \resizebox{\textwidth}{!}{\begin{tabular}{ c c c c c } 
 Method  & 4 & 5 & 6 & 7 \\
 \toprule
 \textbf{\methodname (ours)} & \textbf{0.64} \tiny{$\pm$ 0.01} & \textbf{0.47} \tiny{$\pm$ 0.01} & \textbf{0.49} \tiny{$\pm$ 0.01} & \textbf{0.41} \tiny{$\pm$ 0.01} \\ 
 Rand  & 0.01 \tiny{$\pm$ 0.00} & 0.01 \tiny{$\pm$ 0.00} & 0.00 \tiny{$\pm$ 0.00} & 0.00 \tiny{$\pm$ 0.00} \\ 
 MPC  & 0.00 \tiny{$\pm$ 0.00} & 0.00 \tiny{$\pm$ 0.00} & 0.00 \tiny{$\pm$ 0.00} & 0.00 \tiny{$\pm$ 0.00} \\
 NF  & 0.01 \tiny{$\pm$ 0.00} & 0.01 \tiny{$\pm$ 0.00} & 0.00 \tiny{$\pm$ 0.00} & 0.00 \tiny{$\pm$ 0.00} \\ 
 IQL  & 0.00 \tiny{$\pm$ 0.00} & 0.00 \tiny{$\pm$ 0.00} & 0.00 \tiny{$\pm$ 0.00} & 0.00 \tiny{$\pm$ 0.00} \\ 
 BC & 0.00 \tiny{$\pm$ 0.00} & 0.00 \tiny{$\pm$ 0.00} & 0.00 \tiny{$\pm$ 0.00} & 0.00 \tiny{$\pm$ 0.00} \\ 
\end{tabular}}
    \end{subtable}
    \quad
    \begin{subtable}{.48\linewidth}
      \raggedright
        \small
     \vspace{5pt}
        \caption{\label{tab:block_stacking_combinatorial_generalization_partial} \emph{robogym-rearrange}, partial specification.}
        \vspace{-5pt}
        \resizebox{\textwidth}{!}{\begin{tabular}{ c c c c c } 
 Method  & 4 & 5 & 6 & 7 \\
 \toprule
\textbf{\methodname (ours)} & \textbf{0.47} \tiny{$\pm$ 0.01} & \textbf{0.33} \tiny{$\pm$ 0.01} & \textbf{0.27} \tiny{$\pm$ 0.01} & \textbf{0.22} \tiny{$\pm$ 0.01} \\ 
Rand & 0.005 \tiny{$\pm$ 0.001} & 0.001 \tiny{$\pm$ 0.00} & 0.002 \tiny{$\pm$ 0.001} & 0.001 \tiny{$\pm$ 0.00} \\ 
 MPC & 0.00 \tiny{$\pm$ 0.00} & 0.001 \tiny{$\pm $ 0.001} & 0.00 \tiny{$\pm$ 0.00} & 0.00 \tiny{$\pm$ 0.00} \\
 NF & 0.005 \tiny{$\pm$ 0.001} & 0.001 \tiny{$\pm$ 0.00} & 0.002 \tiny{$\pm$ 0.001} & 0.001 \tiny{$\pm$ 0.00} \\ 
 IQL & 0.00 \tiny{$\pm$ 0.00} & 0.00 \tiny{$\pm$ 0.00} & 0.00 \tiny{$\pm$ 0.00} & 0.00 \tiny{$\pm$ 0.00} \\ 
 BC & 0.00 \tiny{$\pm$ 0.00} & 0.00 \tiny{$\pm$ 0.00} & 0.00 \tiny{$\pm$ 0.00} & 0.00 \tiny{$\pm$ 0.00} \\ 
\end{tabular}}
    \end{subtable} 
    \vspace{-10pt}
\end{table}

\input{src/figs/robogym_tsne}

\subsection{Analysis}
Having quantitatively shown the relative strength of~\methodname in combinatorial generalization from pixels, we now examine how our key design choices of (1) factorizing entity representations into action-invariant and action-dependent features and (2) querying a state transition graph constructed from action-dependent features contribute to~\methodname's behavior and performance.
Is copying the entity type during latent prediction as dSLATE does sufficient for disentangling the location and appearance of objects into the state and type respectively?
Does dSLATE learn to sparsely modify only the entity that corresponds to the moved object in the sensorimotor transition, such that the nodes of the state transition graph meaningfully can be reused across entities?
These are nontrivial capabilities because~\methodname is self-supervised on only the experience buffer.

Fig.~\ref{fig:planning_and_control}b, which visualizes the \texttt{align}, \texttt{select-constraint}, and \texttt{bind} functions of~\methodname on \emph{robogym-rearrange}, suggests that, at least for the simplified setting we consider, the answer to both questions is yes.
\methodname has learned to represent different objects in different slots and construct a graph whose nodes capture location information.
Fig.~\ref{fig:robogym_tsne} shows a t-SNE~\citep{van2008visualizing} plot that clusters entities inferred from the \emph{robogym} environment.
Because we have not provided supervision on what states should represent, we observe there are multiple cluster indices that map onto similar groups of points.
This reveals that multiple different regions of $\mathcal{S}$ appear to be modeling similar states.
We also tried merging redundant clusters, but found that this did not improve quantitative performance.

%% file: src/figs/experiment_tasks.tex
\begin{wrapfigure}[22]{r}{0.25\textwidth}
\vspace{-10pt}
\begin{subfigure}[]{.22\textwidth}
  \centering
    \includegraphics[width=\textwidth]{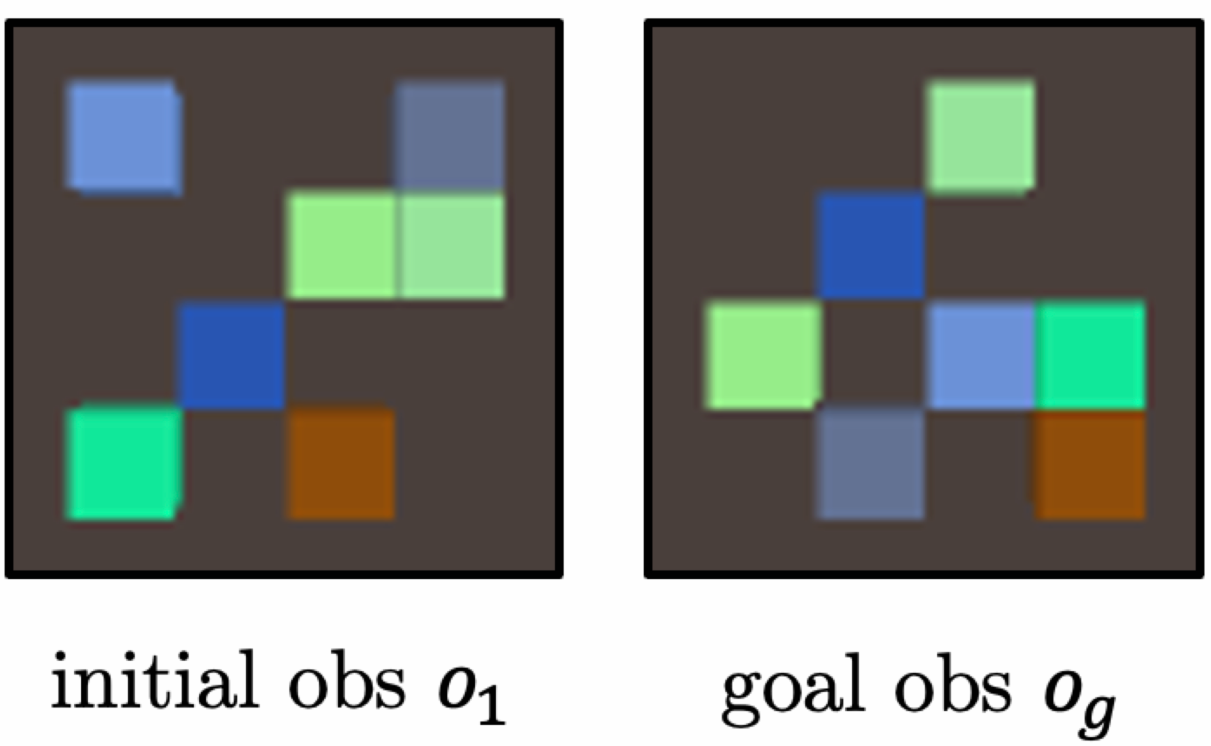}
    \caption{\small{\emph{block-rearrange}}}
    \label{fig:block_rearrange}
\end{subfigure}
\hfill
\begin{subfigure}[]{.22\textwidth}
  \centering
    \includegraphics[width=\textwidth]{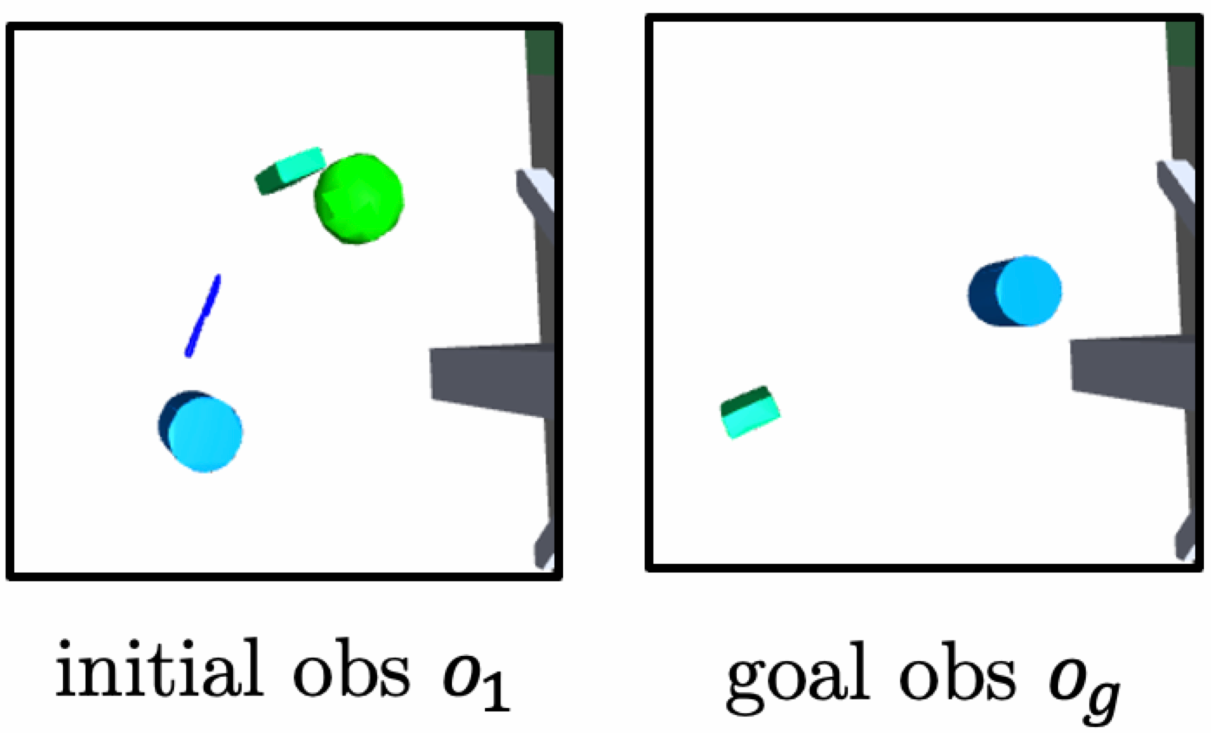}
    \caption{\small{\emph{robogym-rearrange}}}
    \label{fig:robogym_rearrange}
\end{subfigure}
\caption{\small{
Our environments are \emph{block-rearrange} and \emph{robogym-rearrange}.
Fig.~\ref{fig:block_rearrange} shows a complete specification of goal constraints; 
Fig.~\ref{fig:robogym_rearrange} shows a partial specification that only specifies the desired locations for two objects.
}}
\label{fig:environments}
\end{wrapfigure}

%% file: src/figs/robogym_tsne.tex
\begin{wrapfigure}[20]{r}{0.45\textwidth}
  \vspace{-20pt}
  \begin{center}
    \includegraphics[width=0.45\textwidth]{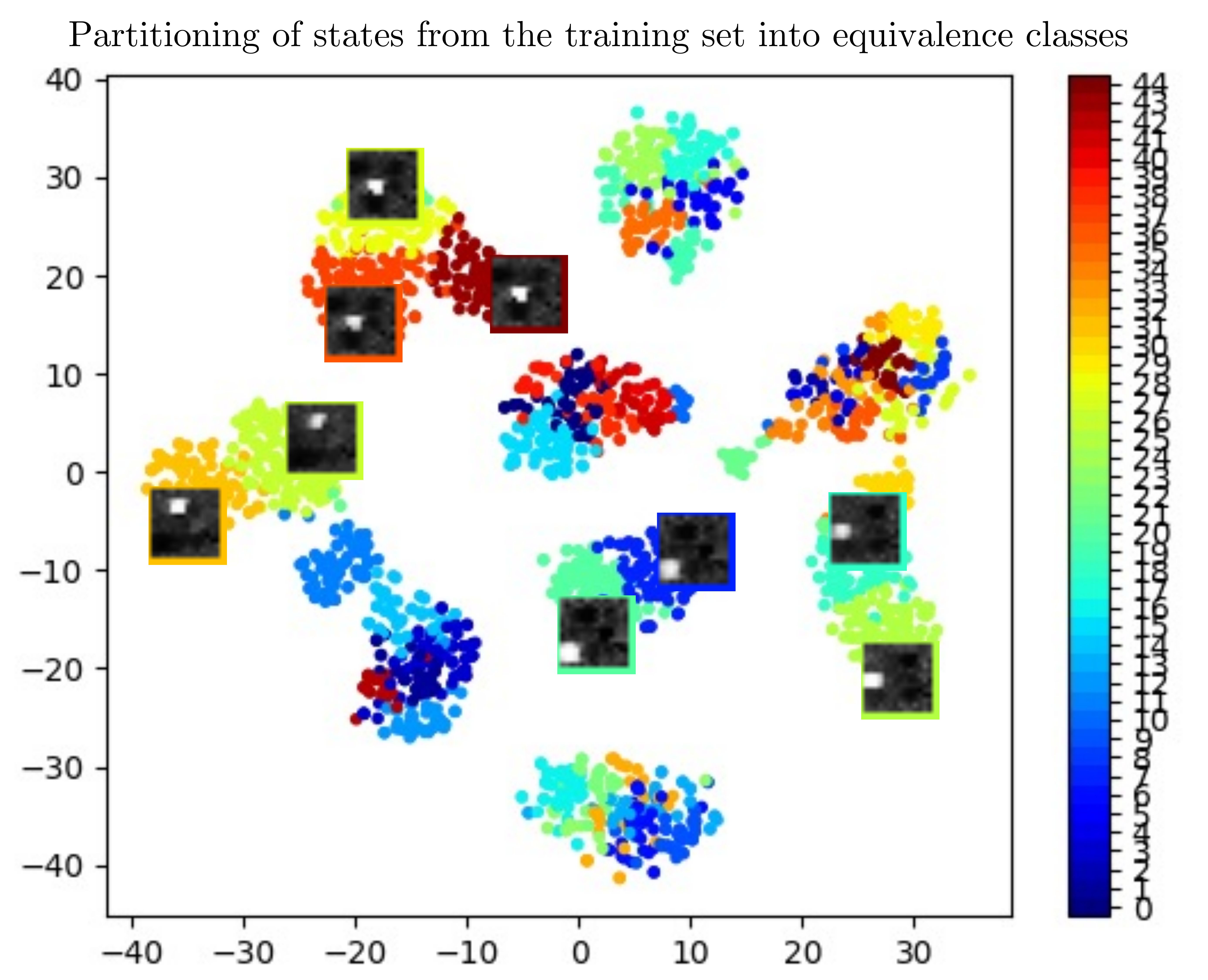}
  \end{center}
  \vspace{-10pt}
  \caption{\small{
  \textbf{Nodes as equivalent classes over states.}
  We show a clustering of states inferred for \emph{robogym-rearrange}, where each cluster centroid is treated as a node in our transition graph.
  We label some clusters with an attention mask computed by averaging the slot attention masks for the entities associated with the cluster.
  }}
  \label{fig:robogym_tsne}
\end{wrapfigure}

%% file: src/discussion.tex
\section{Discussion}
Object rearrangement offers an intuitive setting for studying how an agent can learn reusable abstractions from its sensorimotor experience.
This paper takes a first step toward connecting the world of symbolic planning with human-defined abstractions and the world of representation learning with deep networks by introducing~\methodname.
\methodname is a method for controlling an agent that plans over and acts with state transition graph constructed with entity representations learned from raw sensorimotor transitions, without any other supervision.
We showed that factorizing the entity representation into action-invariant and action-dependent features are important for solving the correspondence and combinatorial problems that make the object rearrangement difficult, and enable~\methodname to significantly outperform existing methods on combinatorial generalization in object rearrangement.
The implementation of~\methodname provides a proof-of-concept for how learning reusable abstractions might be done, which we hope inspires future work to engineer methods like~\methodname for real-world settings.

%% file: src/acknowledgements.tex
\section*{Acknowledgments}
This work was done while MC was an intern at Meta AI.
We would like to thank Leslie Kaelbling for valuable feedback and Yash Sharma and Yilun Du for valuable discussions.
This material is supported in part by the Fannie and John Hertz Foundation, as well as with ONR grant \#N00014-18-1-2873.

%% file: src/appendix.tex
\clearpage

\appendix

\section{Implementation Details}
This section details the implementation design decisions for each component of~\methodname.
The hyperparameters of dSLATE are given in Tab.~\ref{tab:dslate_hyperparameters}.

\subsection{Background: SLATE backbone} \label{appdx:slate_background}
SLATE~\citep{singh2022illiterate} is an autoencoder architecture that uses slot attention (SA)~\citep{locatello2020slot} as a bottleneck.
It preprocesses the image with a discrete variational autoencoder~\citep{ramesh2021zero} into a grid of image features, encodes these features into a grid of tokens, infers slots from this token grid with SA, which also produces an attention mask over the features each slot attends to.
These slots are trained using a transformer decoder~\citep{vaswani2017attention,radford2018improving} to autoregressively reconstruct the tokens using the slots as keys/values.

\begin{table}[h!]
\small
\centering
\begin{tabular}{@{}ll|r@{}}
\toprule

Number of epochs                    &                            & 200      \\
Episodes per epoch                  &                            & 5K      \\
Episode length                      &                            & 5      \\
Batch size                          &                            & 32        \\
Peak LR                             &                            & 0.0002    \\
LR warmup steps                     &                            & 30000     \\
Dropout                             &                            & 0.1       \\ 

\midrule

\multirow{4}{*}{Discrete VAE}       & Vocabulary Size            & 4096      \\
                                    & Temp. Cooldown             & 1.0 to 0.1\\
                                    & Temp. Cooldown Steps       & 30000     \\
                                    & LR (no warmup)             & 0.0003    \\
                                    & Image Size                 & 64       \\
                                    & Image Tokens               & Image Size $/$ 4      \\ 
\midrule

\multirow{3}{*}{
transformer decoder}                & Layers                     & 4         \\
                                    & Heads                      & 4         \\
                                    & Hidden Dim.                & 192       \\ 

\midrule

\multirow{6}{*}{
Slot attention}                     & Slots                      & 5        \\
                                    & Iterations                 & 3         \\
    
                                    & Slot Heads                 & 1         \\
                                    & Slot Dim. ($h$)          & 192       \\
                                    & Type Dim. ($\lambda^z$)      & 96       \\   
                                    & State Dim. ($\lambda^s$)         & 96       \\   
                                                                                               
\midrule
\multirow{3}{*}{
transformer dynamics}               & Layers                     & 4         \\
                                    & Heads                      & 4         \\
                                    & Hidden Dim.                & 96       \\

\midrule
\end{tabular}
\caption{\small{\textbf{Hyperparameters for training dSLATE}
These hyperparameters are almost identical to those found in~\citet[Fig. 7]{singh2022illiterate}, but because dSLATE operates on video demonstrations rather than static images, we changed some hyperparameters to save memory cost.
We changed the batch size from 50 to 32, the number of transformer layers and heads from 8 to 4, the number of slot attention iterations from 7 to 3 without observing a significant change in performance.
Because each video in the experience buffer contains four objects, we used five slots, one more than the number of objects, following the convention used in~\citet{van2018relational,veerapaneni2020entity}. 
}}
\label{tab:dslate_hyperparameters}
\end{table}

\subsection{Constructing nodes by clustering states}
For \emph{block-rearrange}, we found that we obtained better clusterings when we used the SA attention mask $\alpha$ as the state $s$.
For \emph{robogym-rearrange}, we found that we obtained better clusterings when we used the action-dependent part of the SA slot $\bm{\lambda}^s$ as the state $s$.
We also empirically found that certain choices of distance metric used for K-means \texttt{cluster}ing and \texttt{bind}ing (implemented as nearest-neighbors) depended on which choice of state representation we used, and this is summarized in Table~\ref{tab:FACTS_hyperparameters}.
The K-means implmentation is adapted from \url{https://github.com/overshiki/kmeans_pytorch}.

When applying the trained dSLATE to the experience buffer to construct the graph we found that increasing the number of SA iterations improved the entity representations, so even though we trained dSLATE with slot attention three iterations, for constructing the graph we used seven iterations.
Lastly, we found that the number of clusters used to for K-Means is the most important hyperparameter for creating a graph that reflected the state transitions.
We swept over 16 to 50 clusters and report the optimal number of clusters we found in Table~\ref{tab:FACTS_clusters}.

\begin{table}[h!]
\small
\centering
\begin{tabular}{@{}l|cc@{}}
\toprule
State representation                & $\alpha$  & $\lambda^s$ \\
\midrule
\texttt{isolate} distance metric    & cosine  & cosine \\
\texttt{cluster} distance metric     & IoU  & squared Euclidean \\
\texttt{bind} distance metric     & cosine  & squared Euclidean \\
\midrule
\end{tabular}
\caption{\small{\textbf{Hyperparameters for constructing the transition graph with~\methodname}.
This table shows the distance metrics we use for the \texttt{isolate}, \texttt{cluster}, and \texttt{bind} functions described in~\ref{sec:build_graph}.
For \emph{block-rearrange} we use the SA attention mask $\alpha$ as the state $s$, and for \emph{robogym-rearrange} we use the action-dependent part of the SA slot $\bm{\lambda}^s$ as the state $s$.
}}
\label{tab:FACTS_hyperparameters}
\end{table}

\begin{table}[h!]
\small
\centering
\begin{tabular}{@{}l|ccc@{}}
\toprule
                        & \emph{block-rearrange}  & \emph{robogym-rearrange} & \emph{block-stacking} \\
\midrule
number of clusters      & 30  & 45 & 47 \\
\midrule
\end{tabular}
\caption{\small{\textbf{Number of clusters used for constructing the nodes of the transition graph.}
}}
\label{tab:FACTS_clusters}
\end{table}

\subsection{Action selection} \label{appdx:action_selection}

To implement \texttt{align} we use the \texttt{scipy.optimize.linear\_sum\_assignment} implementation of the Hungarian algorithm, with Euclidean distances between the $z^k$'s as the matching cost.

Given the set of current entities $\mathbf{h}_t$ and goal constraints $\mathbf{h}_g$, \texttt{select-constraint} returns the index $k$ of the goal constraint to satisfy next.
By~\methodname' construction, the edge between the nodes that $h^k_t$ and $h^k_g$ are bound to is the state transition that would be executed if the action associated to the edge were taken in the environment.
If~\methodname does not find an edge between the two nodes, such as if $h^k_t$ and $h^k_g$ were incorrectly bound to the graph, then~\methodname simply takes a random action.
texttt{select-constraint} consists of two steps: (1) ranking transitions (2) sampling a transition.

\paragraph{Ranking}
The goal of the ranking step is to compute a ranking among the indices of $(h^1_g, ..., h^K_g)$ to choose which index $k$ to actually select to affect with an action.
Intuitively, we should rank indices $k$ according to how different $s^k_t$ and $s^k_g$ are because a large difference would indicate that the constraint $h^k_g$ is not satisfied, which means we would need to take an action to move the corresponding object represented by $h^k_t$. 
We reuse the distance metric $d(\cdot , \cdot)$ used for \texttt{isolate} to implement this ranking.

\paragraph{Sampling}
Given our ranking, the goal of the sampling step is to select a $k \in \{1, ..., K\}$ whose associated entity we will affect with an action.
One way to do this is to simply choose $k$ as $k = \argmax_{k' \in \{1, ..., \tilde{K}\}} d(s^{k'}_t, s^{k'}_{t+1})$ as in \texttt{isolate}, but we empirically found that sampling $k$ from a categorical distribution whose pre-normalized probabilities are given by $d(s^{k'}_t, s^{k'}_{t+1})$ resulted in better task performance so we used this stochastic sampling approach.
One explanation for why using the argmax may be worse is that it relies on the distance metric $d(\cdot , \cdot)$, and the state representation $s$, to be such that the distance metric flawlessly assigns high value to entities $k$ that need to be moved and low value to entities $k$ that do not need to be moved.
But because the state space $\mathcal{S}$ is learned through the dSLATE training process without explicit supervision on the geometry of the space, a pair of points that should be farther apart than another set of points may not be accurately reflected by using a fixed distance metric $d(\cdot , \cdot)$. 
Future work will investigate imposing explicit supervision on the geometry of $\mathcal{S}$.

\input{src/figs/robogym_env}

\section{Environment Details} \label{appdx:environment_details}

\paragraph{Environments}

\emph{Block-rearrange} is implemented in PyBullet~\citep{coumans2016pybullet} while \emph{robogym-rearrange} is implemented in Mujoco~\citep{todorov2012mujoco}.

\emph{Robogym-rearrange} (see figures \ref{fig:robogym} and \ref{fig:robogym_original}) is adapted from the rearrange environment in OpenAI's Robogym simulation framework ~\citep{robogym2020} and removes the assumption from \emph{block-rearrange} that all objects are the same size, shape, and orientation and the assumption of predefined locations. 
Furthermore, due to 3D perspective, the objects can look slightly different in different locations. 
Objects are uniformly sampled from a set of 94 meshes consisting of the YCB object set \cite{calli2015ycb} and a set of basic geometric shapes, with colors sampled from a set of 13. 
The camera angle is a bird's eye view over the table, and the size of each object is normalized by its longest dimension, so tall thin objects appear smaller. 
The objects' target positions are randomly sampled such that they don't overlap with each other or any of the initial positions, and the target orientation is set to be unchanged.
Because locations take continuous values, we define a match threshold of at most 0.05 for both the initial pick position and the goal placement (the table dimensions are 0.6 by 0.8). 

\paragraph{Sensorimotor interface}
Each observation is a tuple of an initial image displaying the current observation and a goal image displaying constraints to be satisfied -- the goal locations of the objects.
Each action is a tuple $(w, \Delta w)$, where $w$ is a three-dimensional Cartesian coordinate $(x,y,z)$ in the environment arena.
Objects are initialized at random non-overlapping locations that also do not overlap with their goal locations.
For these tasks the $z$ (height) coordinate is always fixed.
An object is picked if $w$ is within a certain threshold of its location.
For \emph{block-rearrange} where object locations are fixed points in a grid, the object is snapped to the nearest grid location to $w+\Delta w$.
Constraints are considered satisfied if objects are placed within a certain threshold of their target location.

\section{Baseline Implementation Details} \label{sec:baseline_implementation_details}

\paragraph{Random (Rand)}
The random policy takes actions using \texttt{env.action\_space.sample()}.

\paragraph{Behavior cloning (BC)} 
This approach trains a policy to output the actions directly taken in the provided dataset. We use an MSE loss to train the policy to imitate the actions.

\paragraph{Implicit Q-learning (IQL)}
IQL is a simple, offline RL approach that uses temporal difference (TD) learning with the dataset actions and  trains a behavior policy value function. To produce an optimal value function, IQL estimates the maximum of the Q-function using expectile regression with an asymmetric MSE using the following objectives:
\begin{align}
    L_V(\psi) &= \mathbb{E}_{(s,a)\sim \mathcal{D}}[L_2^\tau (Q_{\hat{\theta}}(s,a) - V_\psi (s))] \text{ where } L^\tau_2(u)=|\tau - \mathbbm{1}(u<0)|u^2 \\
    L_Q(\theta)&=\mathbb{E}_{(s,a,s')\sim\mathcal{D}}[(r(s,a) + \gamma V_\psi (s') - Q_\theta (s,a))^2] \\
    L_{\pi}(\phi) &= \mathbb{E}_{(s,a)\sim\mathcal{D}}[\exp{(\beta(Q_{\hat{\theta}}(s,a) - V_\psi(s)))} \log{\pi_\phi (a|s)}].
\end{align}
The $V(s)$ estimates are used for TD-backups and the optimal policy is extracted with advantage-weighted behavioral cloning.

\paragraph{Model predictive control (MPC)}
This approach uses model predictive control with the cross entropy method (CEM) to select actions, using the transformer dynamics model of dSLATE to perform rollouts in latent space.
This is similar to the approached used in OP3~\citep{veerapaneni2020entity}, except that we use more recently proposed architectural components (slot attention~\citep{locatello2020slot} instead of IODINE~\citep{greff2019multi}, a transformer instead of a graph network~\citep{battaglia2018relational,van2018relational,chang2016compositional}) so our MPC results are not directly comparable to that of OP3.
We use the same dSLATE checkpoint that was used for~\methodname.

We implement this MPC baseline using the \texttt{mbrl-lib} library~\citep{Pineda2021MBRL} with 10 CEM iterations, an elite ratio of 0.05, and a population size of 250 which was the best configuration we found that fit within a wall clock budget of two days for 8 objects and 100 test episodes.
We swept over CEM iterations of $[5, 10, 20]$, elite ratio of $[0.05, 0.1, 0.2]$, and population sizes of $[250, 500, 1000]$, and found that the elite ratio was the most important hyperparameter.

The cost function is computed by first aligning the predicted slots $\mathbf{h}_T$ and goal constraints $\mathbf{h}_g$ using the same \texttt{align} procedure in Appendx.~\ref{appdx:action_selection}, and then adding up the squared Euclidean distance between slots as $cost = \sum_{k} (h^k_T - h^k_g)^2$.

\paragraph{Non-factorized graph search (NF)}
This approach is an ablation to~\methodname that does not construct a graph over state transitions of individual entities but instead constructs a graph over state transition over entity sets, i.e. each transition is $(\mathbf{s}, a, \mathbf{s}')$ rather than $(s^k, a, s^{k\prime})$.
As with MPC, we use the same dSLATE checkpoint that was used for~\methodname.

The purpose of this ablation is to elucidate the benefit of factorizing the transition graph over \emph{individual entities} rather than \emph{entity sets}.
Because nodes in the transition graph for NF represent a set of entity states rather than individual entity states, we use Dijkstra's algorithm, as in~\citep{eysenbach2019sorb,yang2020plan2vec,zhang2018composable} to plan a unbroken path from the node the initial observation is bound to to the node a goal observation is bound to.
For each time-step, we plan a path along the nodes using Dijkstra's algorithm, then return the action associated with the first edge along that path.
Like~\methodname, NF is a non-parametric model, which means that for a set of entities to be bound to a node in the graph, that node must contain the exact set of entity states corresponding to the states of the entities.
If we do not successfully bind to the graph, or if we do not find a path between the current node and the goal node, we sample a random action as~\methodname does.

\section{Additional Results} \label{appdx:additional_results}
This section presents additional results and analyses of~\methodname.

\subsection{Analysis of key hyperparameters}
In this section, we analyze the sensitivity of task performance to several hyperparameters used in~\methodname when creating the graph: the number of clusters, the number of examples from the experience buffer to use, and the number of slots used in slot attention.
We perform this evaluation in the robogym environment with four objects in the complete goal specification.
As Fig.~\ref{fig:quantitative_analysis} shows, performance depends on the number of initialized clusters and the number of batches from the training set used to construct the graph.
With too few clusters, the clusters are too coarse-grained to differentiate objects in significantly different positions.
With too many, the performance deteriorates as the data is needlessly split into duplicate clusters. 
Performance improves with more data, as the graph has better coverage.
Although~\methodname performs worse when there are insufficient slots to represent all objects present in the environment, performance is barely impacted by having double the number of necessary slots.
Our method can thus still work in environments with an unknown but upper-bounded number of objects.

\input{src/figs/ablations}

\subsection{More computation time for model-based baselines}
We tested whether doubling the computation time for the model-based baselines would improve their performance to be comparable to~\methodname's.
For the results in the main paper, we capped the length of the episode as 4x the minimum number of actions required to solve the task.
In Fig.~\ref{fig:robogym_analysis_baselines}, we vary this interaction horizon multiplier from 1x to 8x.
\methodname degrades less with shorter interaction horizons compared to the baselines.
We find that NF performs similar to the random baseline.
Since NF takes a random action if it cannot bind the given entity set to its graph, this result suggests that the space of subsets of entities is so combinatorially large that NF does not successfully bind to the graph most of the time.
We verified that this is the case by inspecting when NF takes random actions.
MPC performs the worst out of all the methods, performing worse than random.
We tested that the cost function described in Appdx.~\ref{sec:baseline_implementation_details} ranks latents that match the goal constraint with a lower cost than randomly sampled latents, which suggests that the main source of error is due to the inaccuracy in the prediction rollouts.
This can be expected, as learned models suffer from compounding errors when rolled out~\citep{janner2019trust} and prior methods that use MPC for object-centric methods only roll out for very short horizons~\citep{veerapaneni2020entity}.

\input{src/figs/robogym_analysis_baselines}

\subsection{More challenging settings}
Finally, we analyzed~\methodname in more challenging settings that crudely emulate the noisy nature of real-world robotics.
As Fig.~\ref{fig:robogym_analysis} (left) shows,~\methodname is more robust than the baselines to the addition of Gaussian noise to the action at every time step, up until the noise variance is comparable to the maximum distance for successful picking and goal placements. 
The performance remains high given significantly fewer interaction steps (Fig.~\ref{fig:robogym_analysis}, right).
Nevertheless, our success rate is still nowhere perfect, signifying much more work to do in scaling~\methodname to the real world.

\input{src/figs/robogym_analysis}

\section{Combinatorial Space} \label{appdx:combinatorial_space}
This section details the calculation of the combinatorial size of the task space described in \S~\ref{sec:experiments}.
The number of object configurations in the initial state is $|S| \choose k$. 
In the complete specification setting, all objects must be moved, so $t \geq k$.
At each step, any of the $k$ occupied grid cells can be moved to any of the $|S| \choose k$ unoccupied grid cells, so the number of successor states is $k \times (|S| - k)$.
With $|S|$ object locations and $k$ objects, the number of possible trajectories over object configurations of $t$ timesteps is ${|S| \choose k} \times (k \times (|S| - k))^t$.
For \emph{block-rearrange} $|S| = 16$ so with $k=7$ the number of possible trajectories is $\geq 4.5 \times 10^{16}$.

\section{Limitations and future work.} \label{appdx:limitations}
\methodname relies on a nonparametric, non-learning-based approach for control to highlight the generalization capability of our representation of the combinatorial task space, but this limits~\methodname to only composing previously seen transitions for previously seen entities.
Collapsing the combinatorial space along state transitions already provides significant gains but does not adapt to the introduction of novel objects at test time.
\methodname is currently implemented with tools such as SLATE and K-means that have much potential for improvement.
We expect future variations of~\methodname will improve upon our results by replacing SLATE and K-means with their future successors.

Beyond the challenge of improving object-centric models to robustly model real pixels, extending our method to real world environments, such as those studied in~\citet{gokhale2019cooking,chang2020procedure} would require overcoming the additional challenge of translating our high-level pick-and-move action primitives into motor torques for a real robot in a way that handles different object geometries, masses, and properties.
Given that many works in learning robotics (e.g. \citet{devin2020self,yang2021learning}) tackle this exact problem of goal-conditioned object grasping and manipulation, one potential approach to scale our method to real world environments is to train such goal-conditioned policies as the pick-and-move primitives for~\methodname to compose.

In this paper, we have assumed objects can be moved independently.
Preliminary experiments suggest that~\methodname can be augmented to support tasks like block-stacking that involve dependencies among objects, but how to handle these dependencies would warrant a standalone treatment in future work.

\section{Why the name ``Neural Constraint Satisfaction?''} \label{appdx:name}
\methodname can be seen as physically solving a embodied constraint satisfaction problem, where states are variables, identities are variable values, and actions carry out variable assignments.
Unlike symbolic constraint satisfaction, these variables, their domains, the assignment operator, and the constraints are all learned from the sensorimotor interface, hence the name Neural Constraint Satisfaction.

%% file: src/figs/robogym_env.tex
\setlength{\fboxsep}{0pt}
\begin{figure}
    \centering
    \fbox{
    \includegraphics[width=.15\textwidth]{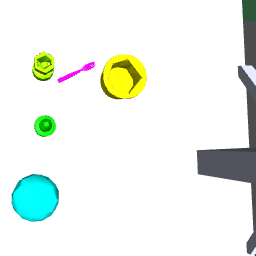}
    }\hspace{-2.5pt}
    \fbox{
    \includegraphics[width=.15\textwidth]{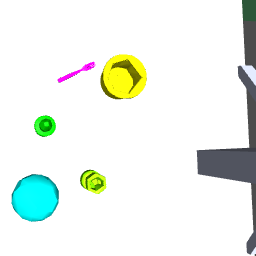}
    }\hspace{-2.5pt}
    \fbox{
    \includegraphics[width=.15\textwidth]{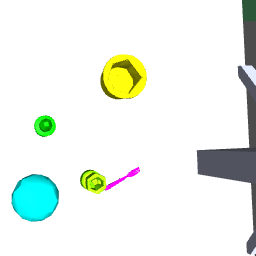}
    }\hspace{-2.5pt}
    \fbox{
    \includegraphics[width=.15\textwidth]{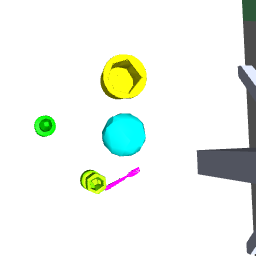}
    }\hspace{-2.5pt}
    \fbox{
    \includegraphics[width=.15\textwidth]{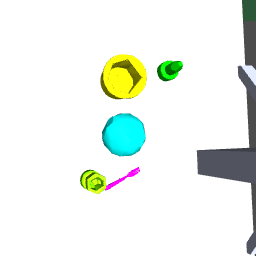}
    }\hspace{-2.5pt}
    \fbox{
    \includegraphics[width=.15\textwidth]{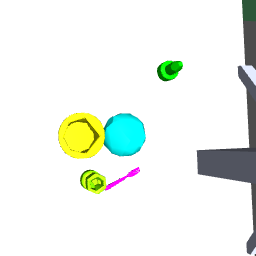}
    }
    \caption{An example of solving a task in the robogym rearrange environment used in this paper. }
    \vspace{-10pt}
    \label{fig:robogym}
\end{figure}
\begin{figure}
    \centering
    \includegraphics[width=0.5\textwidth]{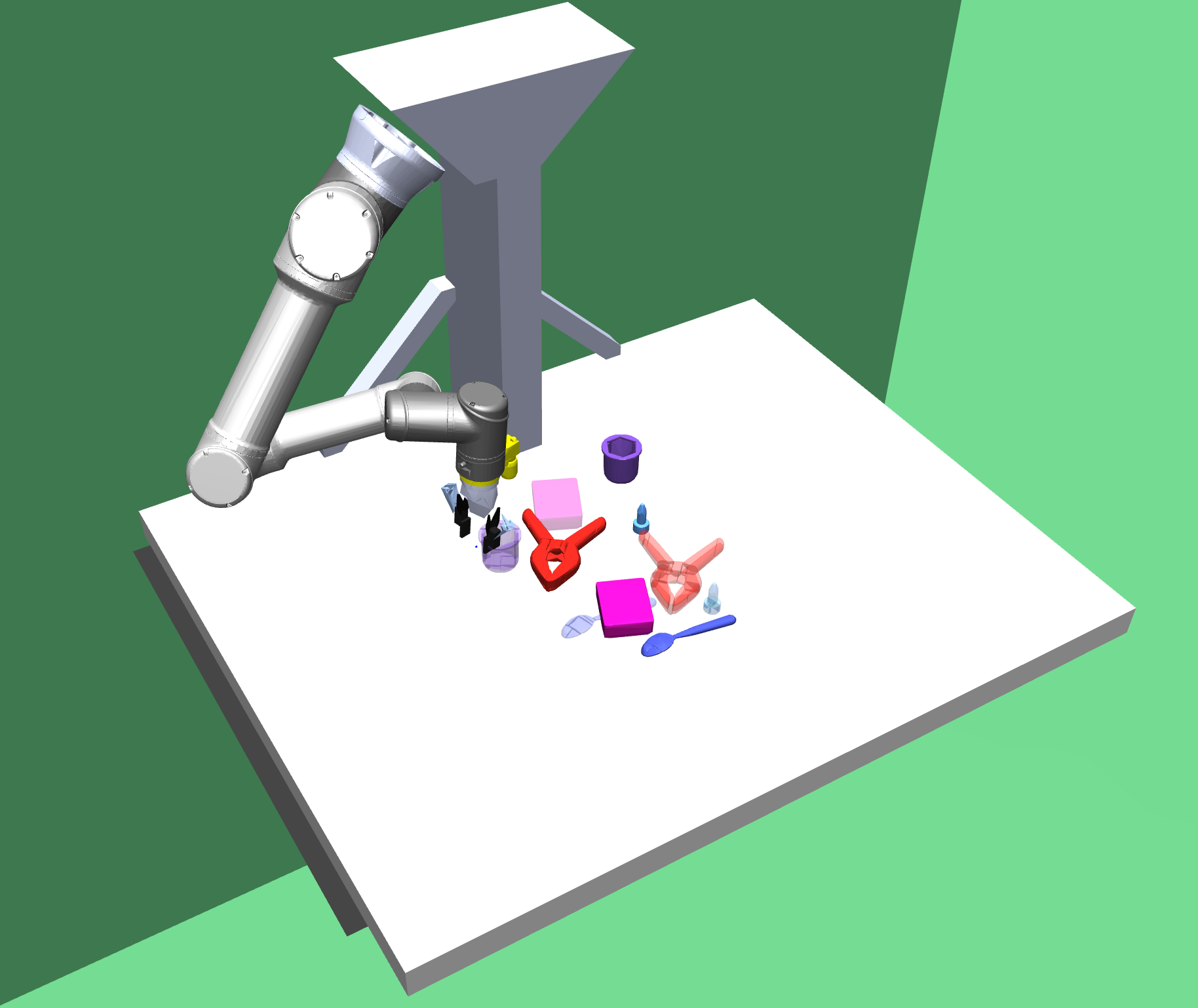}
    \caption{The original Robogym rearrange setup}
    \label{fig:robogym_original}
\end{figure}

%% file: src/figs/ablations.tex
\begin{figure}
    \centering
    \vspace{-10pt}
    \includegraphics[width=.3\textwidth]{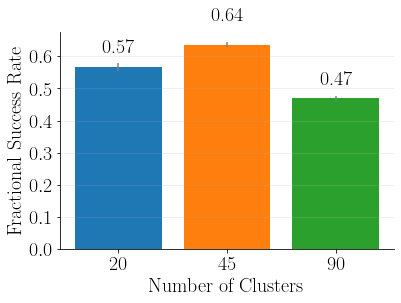}\hspace{10pt}
    \includegraphics[width=.3\textwidth]{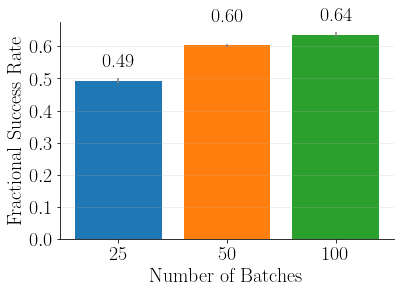}\hspace{10pt}
    \includegraphics[width=.3\textwidth]{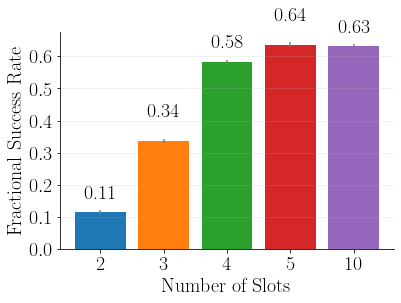}
    \vspace{-5pt}
    \caption{The performance of our method as the number of initialized clusters and batches from the training set used to construct the graph, and the number of slots are varied.}
    \vspace{-10pt}
    \label{fig:quantitative_analysis}
\end{figure}

%% file: src/figs/robogym_analysis_baselines.tex
\begin{figure}[th]
\begin{subfigure}[]{.24\textwidth}
  \centering
    \includegraphics[width=\textwidth]{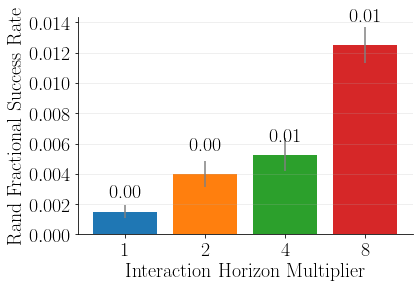}
  \caption{Rand}
  \label{fig:interaction_horizon:rand}
\end{subfigure}
\hfill
\begin{subfigure}[]{.24\textwidth}
  \centering
    \includegraphics[width=\textwidth]{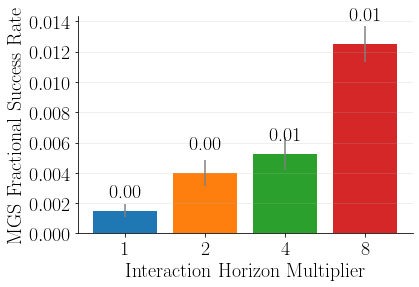}  
  \caption{NF}
  \label{fig:interaction_horizon:NF}
\end{subfigure}
\hfill
\begin{subfigure}[]{.24\textwidth}
  \centering
    \includegraphics[width=\textwidth]{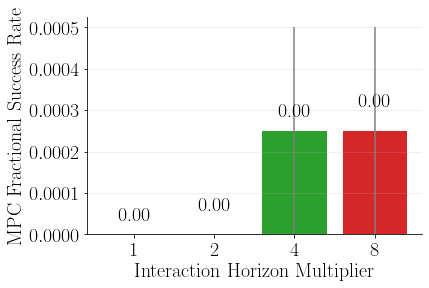}  
  \caption{MPC}
  \label{fig:interaction_horizon:mpc}
\end{subfigure}
\hfill
\begin{subfigure}[]{.24\textwidth}
  \centering
    \includegraphics[width=\textwidth]{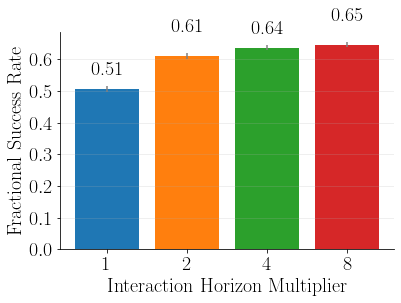}
  \caption{\methodname}
  \label{fig:interaction_horizon:hlg}
\end{subfigure}
\caption{\small{\textbf{Varying interaction horizon.}
The performance of the NF (b) and MPC (c) baselines compared to~\methodname (d, reproduced from Fig.~\ref{fig:robogym_analysis}) and the random baseline (a) on \emph{robogym-rearrange} as we vary the interaction horizon (as a multiple of the minimum steps needed to complete the task).
\emph{Note that the scale of the y-axis is not the same.}
While a longer horizon improves performance,~\methodname still achieves at least 50x better accuracy with an interaction horizon multiplier of 1 than the performance obtained by increasing the interaction horizon multiplier for the model-based baselines to 8.
}}
\label{fig:robogym_analysis_baselines}
\end{figure}

%% file: src/figs/robogym_analysis.tex
\begin{figure}
    \centering
\vspace{-10pt}
    \includegraphics[width=.3\textwidth]{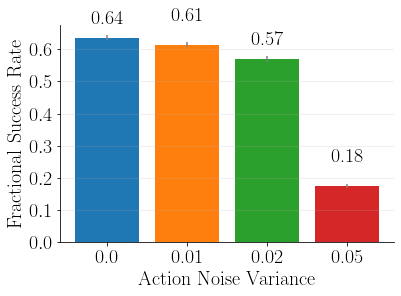}
    \qquad
    \includegraphics[width=.3\textwidth]{figs/horizon.png}
    \caption{\textbf{Stress testing~\methodname}
    This figure shows the performance of~\methodname on \emph{robogym-rearrange} as we vary the amount of noise added to the actions (left) and vary the interaction horizon, defined as a multiple of the minimum steps needed to complete the task (right).}
    \vspace{-10pt}
    \label{fig:robogym_analysis}
\end{figure}